\documentclass{article} 
\usepackage{threeparttable}
\usepackage{iclr2026_conference,times}

\usepackage[utf8]{inputenc} 
\usepackage[T1]{fontenc}    
\usepackage{hyperref}       
\usepackage{url}            
\usepackage{booktabs}       
\usepackage{amsfonts}       
\usepackage{nicefrac}       
\usepackage{microtype}      
\usepackage{xcolor}         
\usepackage{graphicx}
\usepackage{subcaption}  
\usepackage[most]{tcolorbox}

\usepackage{booktabs}
\usepackage{multirow}
\usepackage{float}

\usepackage{indentfirst}

\usepackage{amsmath,amsfonts,bm}

\def\eqref#1{equation~\ref{#1}}

\def\1{\bm{1}}

\DeclareMathAlphabet{\mathsfit}{\encodingdefault}{\sfdefault}{m}{sl}
\SetMathAlphabet{\mathsfit}{bold}{\encodingdefault}{\sfdefault}{bx}{n}

\usepackage{hyperref}
\usepackage{url}

\title{\includegraphics[height=1em]{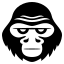}~GUIrilla: A Scalable Framework for Automated Desktop UI Exploration}

\author{Sofiya Garkot, Maksym Shamrai, Ivan Synytsia \& Mariya Hirna \\
MacPaw \\
Kyiv, Ukraine \\
\texttt{\{{sofiyagarkot, mshamrai, sip, maryhirna\}@macpaw.com}
}
}
\iclrfinalcopy 
\begin{document}

\maketitle

\begin{abstract}
The performance and generalization of foundation models for interactive systems critically depend on the availability of large-scale, realistic training data. While recent advances in large language models (LLMs) have improved GUI understanding, progress in desktop automation remains constrained by the scarcity of high-quality, publicly available desktop interaction data, particularly for macOS. We introduce GUIRILLA\footnote{\url{https://github.com/MacPaw/GUIrilla}}, a scalable \emph{data crawling framework} for automated exploration of desktop GUIs. GUIRILLA is not an autonomous agent; instead, it systematically collects realistic interaction traces and accessibility metadata intended to support the training, evaluation, and stabilization of downstream foundation models and GUI agents. The framework targets macOS, a largely underrepresented platform in existing resources, and organizes explored interfaces into hierarchical \emph{MacApp Trees} derived from accessibility states and user actions. As part of this work, we release these MacApp Trees as a reusable structural representation of macOS applications, enabling downstream analysis, retrieval, testing, and future agent training. We additionally release macapptree\footnote{\url{https://github.com/MacPaw/macapptree/}}, an open-source library for reproducible accessibility-driven GUI data collection, along with the full framework implementation to support open research in desktop autonomy.
\end{abstract}

\section{Introduction} \label{sec:intro}

Understanding user interfaces (UI) through machine learning has emerged as a critical challenge in human--computer interaction. Recent advances in large language models (LLMs) have driven rapid progress in multimodal systems that interact with graphical user interfaces, enabling applications ranging from UI automation and assistive technologies to software testing and interactive agents \citep{Kapoor2024OmniACT, qin2025uitarspioneeringautomatedgui, Cheng2024SeeClick, pawlowski2024tinyclicksingleturnagentempowering}. While training models to navigate mobile UIs has been extensively studied \citep{ lee2024exploreselectderiverecall} thanks to the availability of large-scale datasets: RICO\citep{rico}, AITW\citep{Rawles2023Android}, AutoDroid\citep{Wen2023AutoDroid}, progress in desktop environments remains constrained.
 Unlike mobile, desktop environments are cluttered and dynamic: small icon-based controls often encode critical meaning for task execution. Moreover, often users face overlapping windows, popups, dialogs, and system widgets. Among others, the macOS GUI presents particular challenges due to different coordinate systems and UI standards compared to other operating systems. As a result, existing multimodal benchmarks expose three structural flaws that currently limit the construction of scalable, reusable datasets and interaction representations for desktop UIs:

1. \textbf{Manual annotation does not scale.}
    Recent benchmarks \citep{xie2024osworldbenchmarkingmultimodalagents, Kapoor2024OmniACT, Li2025ScreenSpot} rely on human-designed pipelines where tasks, interactions, or UI elements must be manually demonstrated, recorded, and verified. While such supervision is valuable, it is costly and difficult to scale, limiting dataset coverage across applications, UI states, and interaction flows. This bottleneck affects not only agent training, but also downstream applications such as automated UI testing, retrieval, and software analysis.
    
2. \textbf{Single-window UIs misrepresent real usage.}
    Most existing corpora capture clean snapshots of a \emph{single} application window, whereas real desktop usage involves overlapping windows, modal dialogs, background applications, and system widgets. This discrepancy reduces the ecological validity of collected data and limits its applicability for tasks such as automation, testing, trajectory modeling, and long-horizon interaction reasoning across applications.

3. \textbf{Automated collection requires platform-specific design.}
    Constructing diverse, high-quality desktop UI datasets requires OS-specific expertise to handle heterogeneous UI toolkits, event models, and permission systems. Effective automation further demands tailored engineering to reliably parse and interact with each platform. For example, macOS lacks robust virtualization support, significantly complicating large-scale automated crawling compared to Android. As a result, macOS remains severely underrepresented in existing datasets: macOS interfaces comprise only 0.06\% of all samples in \textsc{OS-Atlas} \citep{os-atlas}, and approximately 2.45\% among automatically collected desktop UIs overall.

\begin{figure*}[htbp]
    \centering
    \includegraphics[width=\linewidth]{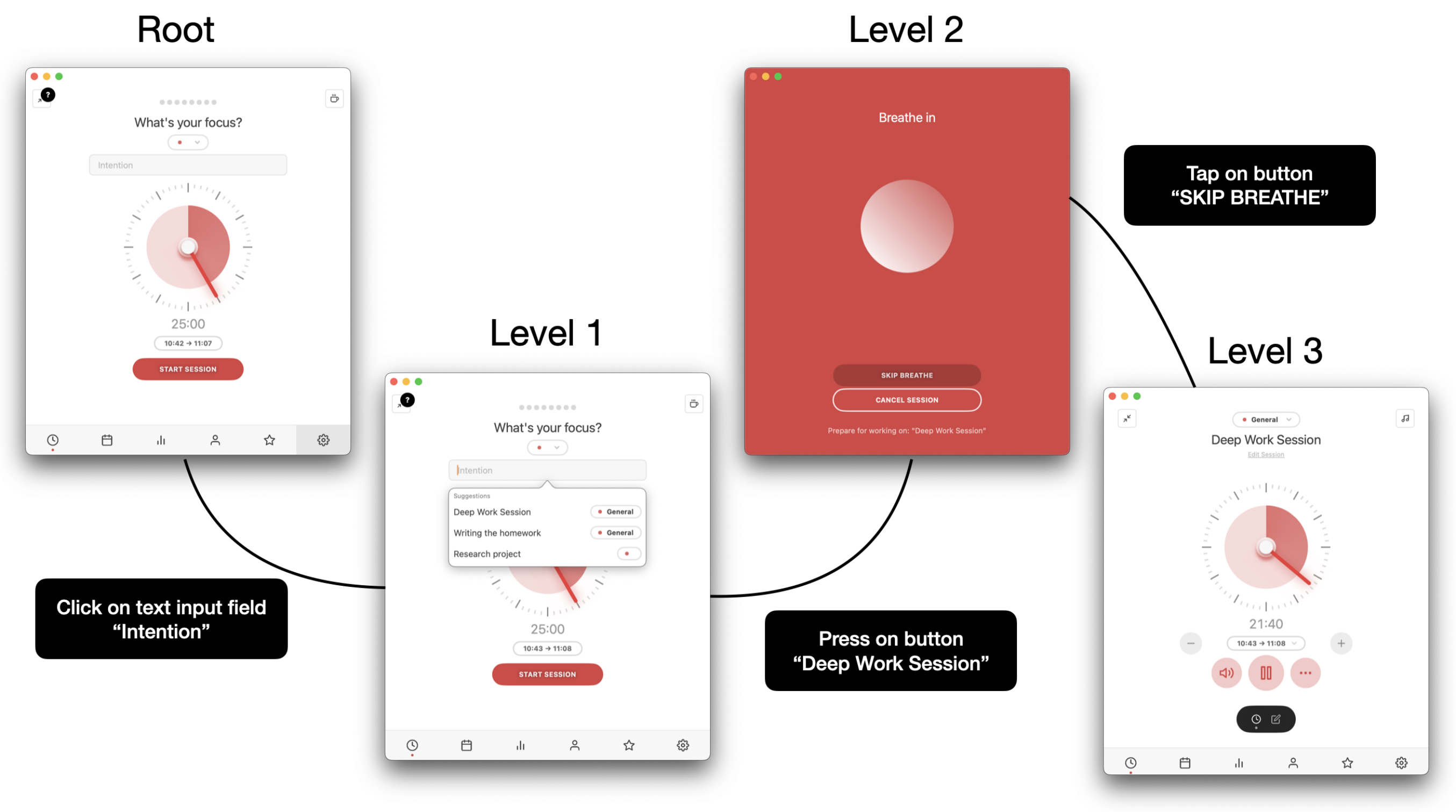}
    \caption{Parsed hierarchical tree structure from the Session application. Each node represents a UI state, containing the full accessibility tree along with a screenshot of the interface, and the edges denote \textsc{GUIrilla} crawler actions. The hierarchy reflects a sequence of interactions as the agent interacts with application UI, forming the application-specific tree.}
    \label{fig:tree-screenshots}
\end{figure*}

Beyond training computer-use agents, large-scale desktop UI interaction data enables a broad range of applications, including automated GUI testing and regression analysis GUI ripping\citep{guiripping2003}, DinoDroid\citep{zhao2022dinodroidtestingandroidapps}, software quality assurance and bug reproduction ReCDroid\citep{recdroid2019}, UI retrieval and search GUing \citep{GUing2024}, and trajectory-based reasoning or retrieval-augmented agents \citep{zhang2025universalretrievalmultimodaltrajectory, Wen2023AutoDroid}. Structured interaction representations have also been explored in early GUI ripping and event-flow graph methods for automated testing, highlighting the long-standing value of modeling UI behavior as graphs rather than isolated screenshots.

To address these gaps, we introduce \textsc{GUIrilla}, a fully automated framework that explores macOS GUIs at scale and summarizes them in a hierarchical \emph{application tree} format (see Figure \ref{fig:tree-screenshots}). Built on macOS' accessibility API, the \textsc{GUIrilla} crawler systematically explores applications through simulated user interactions. Optionally, it can be assisted by LLM--based components for safer element ordering, context-aware input generation, and obstacle handling; however, \textsc{GUIrilla} remains a \emph{data crawler} whose output is intended to train and stabilize downstream models and agents rather than compete with them.

In this work we make the following key contributions:
\begin{itemize}
    \item \textbf{\textsc{GUIrilla} framework.}  
          The first open-source, automated framework tailored for macOS that constructs detailed full-desktop application trees from Accessibility API snapshots and generates function-centric tasks. Application exploration utilizes specialized interaction handlers and can operate both deterministically   
    \item \textbf{GUIrilla MacApp Trees.}  
          We release a large-scale collection of \emph{MacApp Trees}—hierarchical, accessibility-driven representations of macOS applications that capture UI states and user actions across full-desktop environments. These trees provide a reusable structural abstraction of desktop GUI behavior and can be independently used for UI analysis, retrieval, testing, and future agent training.
    
    \item \textbf{\textsc{GUIrilla-Task} dataset.}  
          A macOS, full-desktop corpus of 27,171 tasks across 1,108 applications and ~4.2K unique screens. We also release \textsc{GUIrilla-Gold} (1,283 human-verified tasks) with a 90.26\% human baseline.
    \item \textbf{\textsc{GUIrilla-See} vision--language models.} 
      We release three models: \textsc{GUIrilla-See} (0.7B), \textsc{GUIrilla-See} (3B), and \textsc{GUIrilla-See} (7B), trained on just 4.2K unique macOS full-desktop screens from \textsc{GUIrilla-Task}. The models serve as a concrete demonstration that realistic macOS coverage and function-level supervision can yield strong transfer to downstream GUI understanding benchmarks with substantially less training data than large multi-OS synthetic corpora.
    \item \textbf{Open-source reproducible toolkit.}  
         Complete end-to-end implementation including data generation pipeline, model training code, evaluation framework, and the macapptree library for collecting accessibility metadata and screenshots, facilitating reproducible automated data collection efforts on macOS.
\end{itemize}

\section{Related work}
\label{sec:rel}

UI understanding has seen rapid progress in \textit{mobile} \citep{Rawles2023Android, Rawles2024AndroidWorld} and \textit{web} \citep{liu2024visualwebbenchfarmultimodalllms, liu2024harnessingwebpageuistextrich} domains, driven largely by the availability of structured interface representations (e.g., HTML/XML) and large-scale datasets such as \textsc{RICO}. These resources enable models to jointly reason about visual appearance, textual content, and interaction affordances at scale. In contrast, desktop environments lack a unified structural representation, exhibit greater visual and behavioral heterogeneity, and often require system-level permissions for interaction and inspection. These properties make scalable, automated data collection substantially more challenging.

Several recent efforts have focused on constructing datasets and benchmarks for desktop UI understanding. ScreenSpot~\citet{Cheng2024SeeClick} and its extensions ScreenSpot-v2 and ScreenSpot-Pro~\citet{Li2025ScreenSpot} introduce task-oriented datasets for evaluating element localization and instruction following on desktop screenshots. Related benchmarks such as WinClick~\citep{WinClick2025}, UI-Vision~\citep{nayak2025uivisiondesktopcentricguibenchmark}, GUI-360$^\circ$~\citep{mu2025gui360circcomprehensivedatasetbenchmark}, and MMBench-GUI~\citep{wang2025mmbenchguihierarchicalmultiplatformevaluation} further expand evaluation coverage across platforms, resolutions, and task granularities. While these benchmarks have advanced evaluation methodology, they typically rely on manually curated task definitions or limited interaction traces, and do not expose the underlying processes used to explore applications or construct interaction trajectories.

Beyond benchmarking, several works explore automated data collection for desktop GUIs. OmniACT~\citep{Kapoor2024OmniACT} presents a manually collected multi-platform dataset spanning macOS, Linux, and Windows, but is limited in scale and application coverage. OS-Atlas~\citep{os-atlas} automates macOS data collection via the Accessibility API, producing large numbers of single-step question--answer pairs. However, its exploration strategies are shallow, rely on raw accessibility labels, and—crucially—the end-to-end crawler implementation is not publicly released, limiting reproducibility and reuse. More recent pipelines such as pyautogui, UI-E2I-Synth~\citep{liu2025uie2isynthadvancingguigrounding}, and OS-Genesis~\citep{os-genesis} leverage LLMs to annotate UI elements or synthesize action trajectories at scale, but focus primarily on generating supervision for grounding or agent training rather than releasing reusable structural representations of application behavior.

A substantial body of work targets GUI grounding and parsing, including OmniParser~\citep{lu2024omniparserpurevisionbased}, HyperClick~\citep{zhang2025hyperclickadvancingreliablegui}, LASER~\citep{wang2025learningactiveperceptionselfevolving}, Aria-UI~\citep{yang2025ariauivisualgroundinggui}, and related methods that improve element localization, uncertainty estimation, or active perception. These approaches primarily study model architectures, training strategies, or supervision signals for grounding benchmarks. While complementary to our work, grounding-focused methods typically assume the availability of curated datasets and do not address the upstream problem of scalable, reproducible desktop UI exploration and representation.

Modeling UI behavior as structured graphs or state-transition systems has a long history in software engineering and automated testing. Early GUI ripping and event-flow graph methods~\citep{guiripping2003} aimed to reverse-engineer application behavior to generate test cases. Subsequent work on Android UI testing and QA—including ~\citep{yoon2023autonomouslargelanguagemodel}, ReCDroid, DinoDroid, and model-based testing frameworks—demonstrated the value of automated exploration for software reliability and regression analysis. More recently, trajectory-based reasoning and retrieval methods have been explored for agentic systems and retrieval-augmented interaction modeling. These lines of work highlight the importance of capturing \emph{interaction structure}, not just isolated screenshots or annotations.

In contrast to prior efforts, \textsc{GUIrilla} focuses on the scalable, automated construction of \emph{MacApp Trees}—hierarchical, accessibility-driven representations that capture UI states and user actions across full-desktop macOS environments. Our framework emphasizes reproducible exploration, safe interaction handling, and the release of reusable structural artifacts rather than solely task annotations or benchmark scores. The resulting trees and derived datasets can support a broad range of downstream applications, including UI automation, testing, retrieval, accessibility analysis, and future agent training.

\section{Methodology}\label{sec:methods}

\textsc{GUIrilla} introduces a \emph{tree-centric, fullscreen} exploration pipeline for macOS GUIs. Our framework builds on macOS Accessibility API\footnote{\url{https://developer.apple.com/documentation/accessibility/accessibility-api}}, while integrating three specialized agents that interpret accessibility metadata, prioritize interface elements, and generate contextually appropriate actions. 

\subsection{macapptree}
To navigate between interface states, \textsc{GUIrilla} leverages the macOS Accessibility API. However, interfacing with this API directly via Python is often cumbersome. To solve this, we developed and open-sourced \emph{macapptree}: a Python package designed to extract an application’s accessibility tree and serialize it into a clean, hierarchical JSON format.

As demonstrated in Table~\ref{tab:sspro_results}, Visual Language Models (VLMs) often struggle with precision on UI benchmarks, leading to navigation errors. By utilizing text-based accessibility trees instead of raw screenshots, our approach allows the use of standard Large Language Models (LLMs). This methodology ensures that as long as the application's accessibility metadata is accurate, the agent interacts only with explicitly defined UI elements --- guaranteeing that the selected coordinates for actions like clicking are always precise.

Beyond its role in our pipeline, \emph{macapptree} could serve in other use cases:
\begin{enumerate} 
    \item \textbf{Accessibility Testing:} Validating whether UI elements are correctly exposed to assistive technologies. 
    \item \textbf{UI Automation:} Providing a reliable, non-visual interface for cross-app interactions. 
    \item \textbf{Visual Debugging:} Allowing developers to inspect exactly how the system interprets the screen. 
\end{enumerate}

This versatility makes the tool valuable for both academic research and practical software engineering workflows.

\subsection{Crawler}

\begin{figure*}[htbp]
    \centering
    \includegraphics[width=\linewidth]{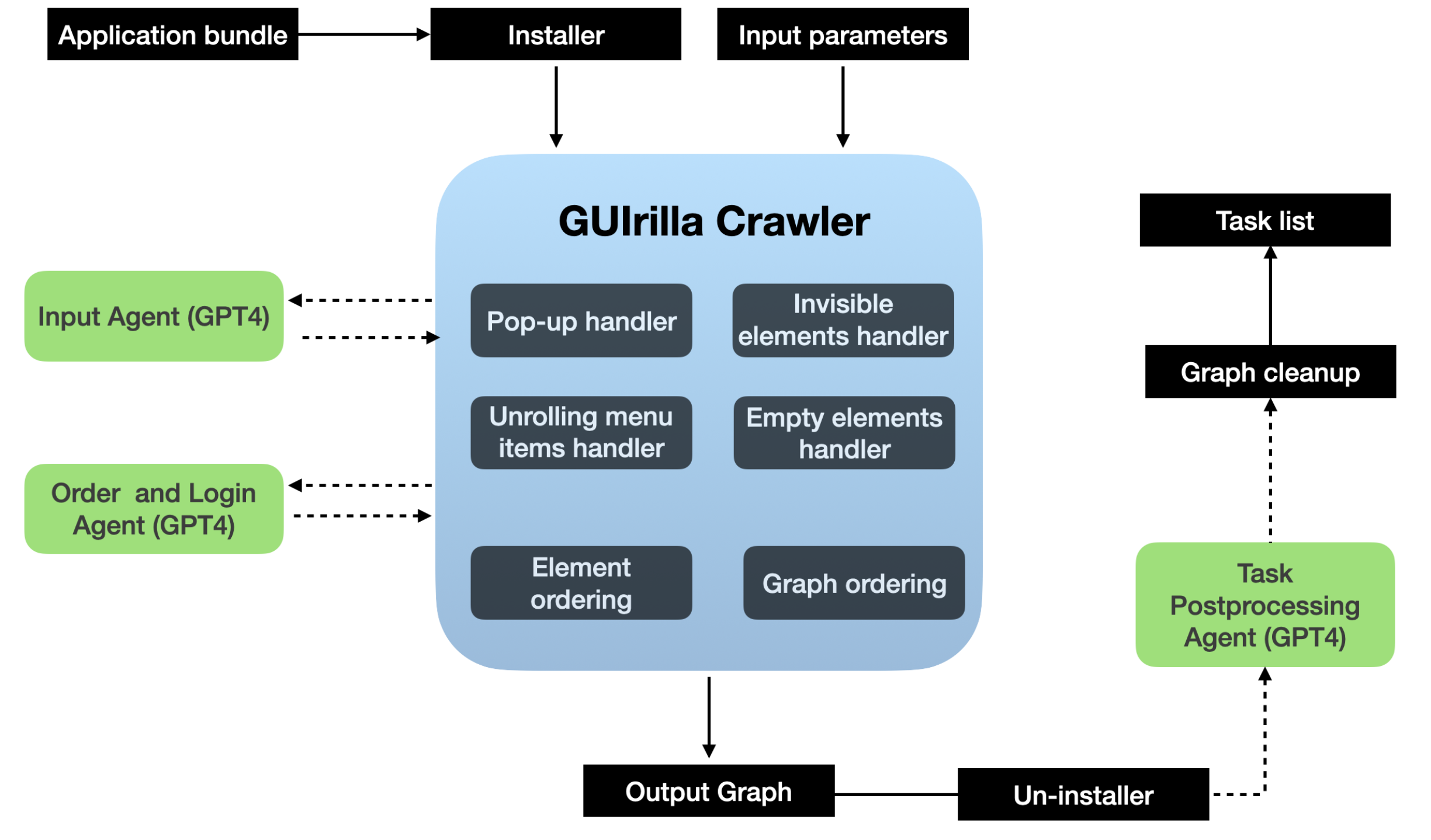}
    \caption{Architecture of the \textsc{GUIrilla} framework. The \textsc{GUIrilla} crawler, equipped with various UI handlers, processes an application bundle using input parameters and installer routines. It interacts with autonomous GPT-4 agents (Input, Order, and Login Agents) to navigate the application. The resulting output tree is refined by a Task Postprocessing Agent (GPT-4), which handles uninstallation and tree cleanup, ultimately producing a structured task list. The dashed line denotes the optional usage of LLMs for app exploration.}
    \label{fig:crawler}
\end{figure*}

The single-app processing pipeline is illustrated in Figure~\ref{fig:crawler} and has the following stages. First, the input bundle undergoes a standard installation routine, and together with user-specified set of parameters (such as maximal desired tree depth, and duration of parsing, the full list is available in Appendix \ref{appendix:parameters}), crawler manages each of the windows of the installed app. Upon installation, the crawler attempts to extract an application's accessibility tree according to macOS accessibility framework. This framework enables simpler interaction with  UI elements on the screen grouping them into a hierarchical tree structure, where each element contains essential properties such as name, role, description, position, and size.
However, application developers must manually annotate or update accessibility metadata.
This manual annotation process often results in error-prone accessibility trees with significant limitations: some trees contain UI elements that remain in the tree after disappearing from view, others include components with incorrect role classifications, and inaccurate positioning information.

To handle these edge cases, our \textsc{GUIrilla} crawler incorporates multiple specialized handlers, as shown in Figure \ref{fig:crawler}.
Within the crawler's core (highlighted in blue), multiple handlers address the common parsing challenges: Pop-up handler manages transient modal content, Invisible elements handler filters off-screen components present in accessibility, Unrolling menu items handler processes dynamically generated navigation elements, and Empty elements handler resolves placeholder elements with missing metadata. This multi-handler approach enables robust extraction of actionable interface information despite the underlying data quality issues.

The \textsc{GUIrilla} crawler performs three types of interactions to explore an application: click, cursor move, type, and press Enter key using \textit{pyautogui}~\citep{pyautogui} library. To enable meaningful interaction with applications, the parsing is supported by three GPT4-based agents (the prompts are available in Appendix \ref{appendix:prompts}):

1. The \textit{Input Agent}: This agent generates contextually appropriate input strings based on the accessibility tree, ensuring relevant text is entered into form fields and search boxes. 

2. The \textit{Order and Login Agent}: Given a hierarchical list of on-screen elements, the agent determines an safest interaction sequence starting with elements that cause minimal UI changes and progressing to those with potentially significant effects (e.g., "Delete" buttons). Login pages are treated as a special case, requiring human input. This agent enhances the security and safety of the exploration process by avoiding random or destructive actions. 

3. The \textit{Task Agents}: After the uninstallation phase, these agents refine the resulting output tree, cleaning up duplicates, and transforming the structured data into a readable list of natural language tasks. Their inclusion enables both refinement and generation of more complex and natural language task descriptions.

While our framework leverages GPT-based agents to enable robust and secure interaction, both the application trees and task data can also be collected deterministically without GPT-4 requests by following a fixed element processing order and using default input string values. However, incorporating GPT-based reasoning significantly improves the safety and contextual relevance of interactions. A detailed comparison between deterministic and GPT-guided exploration is provided in Appendix~\ref{appendix:gpt4-support}.

\subsection{Tree Structure}

The application tree collected with \textsc{GUIrilla} crawler consists of nodes and edges that represent application states and actions, respectively (see Figure~\ref{fig:tree-screenshots}). All interaction trees are automatically annotated and visualized as accompanying SVG files. Across applications, the trees have an average depth of 3.5, with the deepest tree reaching a depth of 101. Each node corresponds to a specific UI state of an application and contains the following fields: 
{\setlength{\itemsep}{2pt}%
\setlength{\parskip}{0pt}%
\begin{itemize}
    \item \textit{Element}: The accessibility tree of the application window at a state.
    \item \textit{Image name}: The filename of the full desktop screenshot associated with a state. 
    \item \textit{Actions}: A list of actions that can be executed without causing significant changes to the UI. We define a significant change as the addition or removal of more than 10 UI elements following an interaction.
\end{itemize}}

Each edge captures a possible interaction and includes:

{\setlength{\itemsep}{2pt}%
\setlength{\parskip}{0pt}%
\begin{itemize}
    \item \textit{Action}: Information about the UI element that triggered the interaction, along with a human-readable action description and a structured dictionary representation that has a 1-to-1 map to \textit{pyautogui} commands. 
    \item \textit{Out vertex}: The resulting UI state after the interaction of the crawler with the GUI.
\end{itemize}}

\begin{figure*}[htbp]
    \centering
    \includegraphics[width=\linewidth]{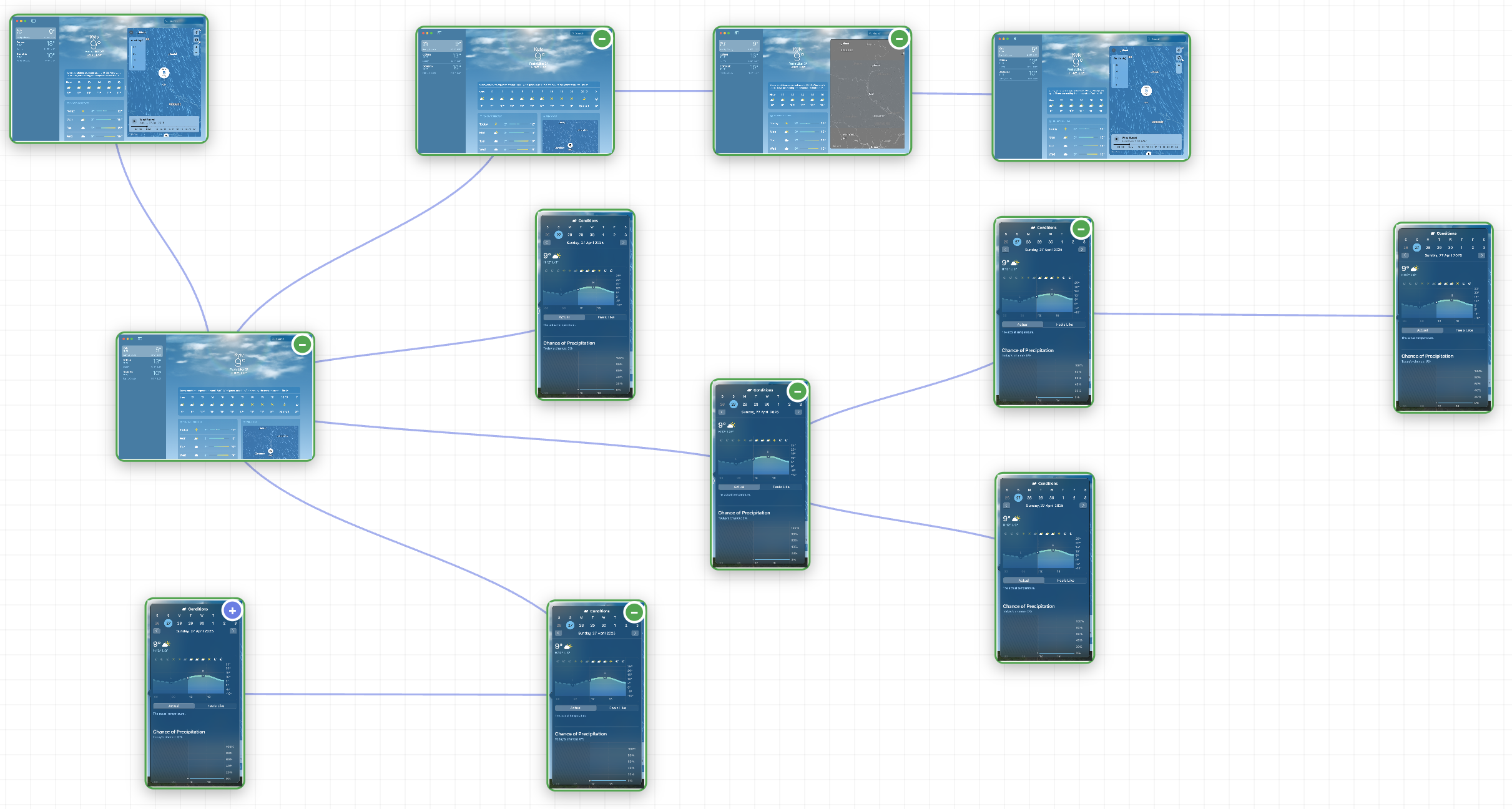}
    \caption{Example of trees based on \textsc{GUIrilla} crawler}
    \label{fig:trees}
\end{figure*}

We release all raw trees as open-source artifacts for broader reuse across research and engineering tasks which includes trees and screenshots for different applications. In total, the dataset comprises 561 GB of compressed data. The examples of tree are provided in Figure \ref{fig:trees}.

\section{Results}
\label{sec:results}

\subsection{Derived Task Dataset from MacApp Trees}

As an application of the collected MacApp Trees, we derive a task-oriented dataset, \textsc{GUIrilla-Task}, by extracting function-centric interactions from tree edges. We deployed the \textsc{GUIrilla} crawler on 12,298 macOS applications using an open Mac App Store dataset~\citep{BibEntry2025May}. Out of these, 1,108 applications supported the macOS Accessibility framework and yielded valid interaction trees.

The resulting task dataset spans a wide range of domains, including productivity, creative tools, system utilities, and developer software, ensuring diverse coverage of real-world desktop UI paradigms. In total, \textsc{GUIrilla-Task} contains 27,171 tasks across 23 application genres, derived from approximately 4.2K unique full-desktop screens.

Each task corresponds to a concrete user interaction (mouse click or keyboard input) paired with a full-desktop screenshot and the associated accessibility tree state. Tasks range from simple navigational actions (e.g., “open settings”) to higher-level functional instructions (e.g., “change your working hours to end at 18:00”). Tasks are annotated with task type (e.g., navigation, settings) and target element category (e.g., button, menu, input field).

Additional dataset statistics, representative samples, and entry attributes are provided in Appendix~\ref{appendix:dataset-stats}. Data collection was performed on a cluster of four M1 Mac Mini machines (16\,GB RAM) running macOS 14.7.5, as well as two MacBook Pros, with multiple parallel exploration environments per host.

\subsection{Downstream Agent Evaluation}

\paragraph{Models.} To illustrate the utility of \textsc{GUIrilla}-derived data for downstream applications, we evaluate a range of vision--language models (VLMs) on tasks extracted from MacApp Trees. These include proprietary systems like OpenAI Computer Use~\citep{openai2025cua} and Claude Computer Use~\citep{anthropic2024computeruse} as well as open-source models: UI-TARS 1.5 (7B), UI-TARS (2B)~\citep{qin2025uitarspioneeringautomatedgui}, Qwen 2.5 VL (7B, 3B)~\citep{bai2025qwen2}, CogAgent 9B~\citep{Hong2023CogAgent}, and OS-Atlas Pro 7B~\citep{os-atlas}.

\paragraph{Metrics.} We report task success rates based on action accuracy. For click tasks, success requires the predicted coordinates to fall within the target element's bounding box. Input tasks additionally require exact text matches.

\paragraph{Results.} Without task-specific fine-tuning, models struggle particularly with text-input interactions, reflecting the complexity of long-horizon desktop behavior. Proprietary systems achieve the strongest performance overall, while open-source models benefit substantially from training on realistic macOS interaction data. Full quantitative results are reported in Appendix~\ref{appendix:eval-agentic}.

\subsection{Ablation study}

\paragraph{Impact of Accessibility Handlers on Exploration Coverage.} Native accessibility annotations vary inconsistently across applications, creating barriers to systematic exploration. The accessibility handlers anticipate UI changes and execute meaningful interactions beyond basic clicking.
Testing on three macOS applications (Stocks, Maps, Weather) across tree depth, duplicate rate, task diversity, and process time shows handlers increase task discovery by 5× in Stocks and 3× in Maps while reducing duplicates and processing time (Appendix~\ref{appendix:ablation-handlers}). These handlers target platform-agnostic problems: inconsistent element labeling, hidden components, and dynamic content. The logic transfers to other operating systems facing similar accessibility inconsistencies.

\paragraph{Generative Task Agents.} We compare two training approaches: (1) deterministic accessibility metadata (\texttt{name}, \texttt{role}, \texttt{role\_description}, \texttt{value}), and (2) GPT-4 task descriptions from screenshots and element crops (Table \ref{tab:task_examples-ablation}).
Accessibility metadata often reduces to generic labels like "button" or "text" without capturing functional intent. In contrast, GPT-generated descriptions understand visual context and explicit purpose. When UI screens contain similar elements, accessibility labels create ambiguous supervision signals that hinder target identification. Florence-0.7B achieved 53.55\% accuracy on GPT-generated tasks versus 40.35\% on accessibility-based tasks—a 13-point gap demonstrating that functional supervision outperforms surface-level properties for training UI agents.

\section{Impact, Limitations, and Ethics}

\paragraph{Broader Impact.} This work has significant potential to advance accessibility technology development, directly benefiting users with disabilities who rely on assistive technologies. By systematically collecting UI interaction data, our framework can improve screen readers, voice-controlled interfaces, and other adaptive technologies that help users navigate complex desktop environments. 

\paragraph{Technical Limitations.} Our approach is primarily constrained by dependence on developer-provided accessibility metadata, which exhibits considerable variation in quality across applications. While currently implemented for macOS, the methodology can be adapted to other platforms such as Windows\footnote{\url{https://learn.microsoft.com/en-us/dotnet/framework/ui-automation/}}, Linux\footnote{\url{https://gnome.pages.gitlab.gnome.org/at-spi2-core/libatspi/}}, and Android\footnote{\url{https://developer.android.com/reference/android/view/accessibility/AccessibilityNodeInfo}} by leveraging their existing accessibility infrastructures, though this requires platform-specific engineering. Additionally, solutions like OmniParser or Screen2AX \citep{muryn2025screen2ax} can be used to remove full reliance on accessibility metadata.

\subsection{Ethical Considerations}
We acknowledge potential risks including privacy violations, security circumvention, and malicious automation. To mitigate these concerns, we implement technical safeguards:
\begin{itemize}
    \item \textbf{Sandboxed Environments}: We strongly recommend conducting data collection in dedicated environments with anonymized profiles to prevent accidental data leakage
    \item \textbf{Local-Only Operation}: All collection, replay, and annotation occur entirely locally without requiring data transmission to third parties
    \item \textbf{Deterministic Handlers}: Rule-based handlers enable fully offline, privacy-preserving automation without external API dependencies
    \item \textbf{Limited API Access}: Framework operates strictly via public macOS Accessibility APIs with no privileged system calls
    \item \textbf{Security-Critical Exclusion}: We explicitly avoid interaction with authentication, payment, or CAPTCHA-related interfaces
\end{itemize}

\paragraph{Responsible Use Guidelines.}
We explicitly discourage malicious use through clear documentation and recommend: (1) running crawlers only in controlled environments with synthetic inputs, (2) applying data filtering to remove sensitive content, and (3) using deterministic handlers for regulated data. Acceptable use cases include academic Human-Computer-Interaction research, accessibility technology development, and educational applications in controlled environments. Prohibited uses include automation of financial/healthcare systems, security circumvention, unauthorized personal data collection, and creation of tools for harassment or illegal activities. We remain committed to community oversight and transparent release practices, maintaining openness to policy revisions based on feedback to ensure responsible deployment of UI automation capabilities.

\paragraph{Use of Large Language Models.} Portions of this manuscript were refined with the assistance of large language models (LLMs) for grammar and style.

\section{Conclusions and Future Directions}

We introduce \textsc{GUIrilla}, a fully automated framework that addresses the data scarcity challenge in desktop UI research by enabling scalable, reproducible exploration of macOS applications. By leveraging accessibility-driven crawling and structured interaction handling, \textsc{GUIrilla} constructs hierarchical MacApp Trees that capture UI states and user actions across full-desktop, multi-application environments.

Unlike prior work that emphasizes task annotation or benchmark performance, our contribution centers on releasing reusable structural representations of desktop UI behavior, along with an open-source crawling framework and tooling. These artifacts support a wide range of downstream applications, including UI automation, software testing, accessibility analysis, retrieval, and agent training, and can be reused independently of any specific model architecture.

\paragraph{Future Work.} While \textsc{GUIrilla} currently leverages native accessibility APIs on macOS, future work could integrate image-to-accessibility or vision-based UI parsing techniques to enable crawling in environments where accessibility metadata is incomplete or unavailable. Another promising direction is the development of local vision--language model (VLM) agents that actively explore applications using reinforcement learning or related approaches, allowing more adaptive exploration strategies and improved coverage of complex interaction patterns. Beyond macOS, the framework can be extended to additional operating systems by adapting platform-specific accessibility interfaces, and its continuous crawling pipeline enables long-term data collection as applications evolve. We hope this work lays the foundation for future research on structured desktop UI representations and serves as an open, extensible resource for the community to build upon as interactive systems continue to advance.

\subsubsection*{Acknowledgments}
We thank the Armed Forces of Ukraine for providing security to complete this work.
We thank Yaroslav Tereshchenko and Victor Muryn for their support, insightful discussions, and assistance in framing the paper and releasing the dataset as open source.

\bibliography{iclr2026_conference}
\bibliographystyle{iclr2026_conference}

\appendix
\newpage
\section{Appendix}
\subsection{Dataset statistics}\label{appendix:dataset-stats}

The collected tasks were split into train and test subsets, such that the applications in test did not appear in train, and test applications contained larger, more complicated accessibility trees. There are $881$ applications with $25,606$ entries in train and $227$ applications with $1,565$ task entries in test.

\subsubsection{Representative sample from the dataset}

Each sample in the dataset includes the following structured fields:
\begin{itemize}
    \item \textbf{Screen ID}: Unique identifier for the UI screen.
    \item \textbf{App Name}: Bundle identifier of the application.
    \item \textbf{Task}: Natural language description of the agent's objective.
    \item \textbf{Raw Action}: Deterministic textual representation of the user action.
    \item \textbf{Action}: Structured action format, e.g., \texttt{"left click, (x, y)"}.
    \item \textbf{Element Data}: JSON metadata of the target UI element extracted from the accessibility tree.
    \item \textbf{Scaling Factor}: Display scaling factor for the specific screen.
    \item \textbf{Original Task}: Boolean indicating whether the task was derived directly from the original interaction tree.
    \item \textbf{A11y Path}: Full accessibility tree before the action was taken.
    \item \textbf{Image}: Full-screen desktop screenshot, stored as a binary image.
    \item \textbf{Cropped Image}: Subregion of the full screenshot containing the target application (variable dimensions).
    \item \textbf{Segmented Image}: Screenshot of the application window with segmented UI regions.
    \item \textbf{Task Category}: One of 22 predefined task categories (e.g., \textit{Search \& Information}, \textit{Files}).
    \item \textbf{Element Category}: One of 16 UI element types (e.g., \textit{Slider}, \textit{Button}).
\end{itemize}

\begin{figure}[H]
    \centering
    \includegraphics[width=\linewidth]{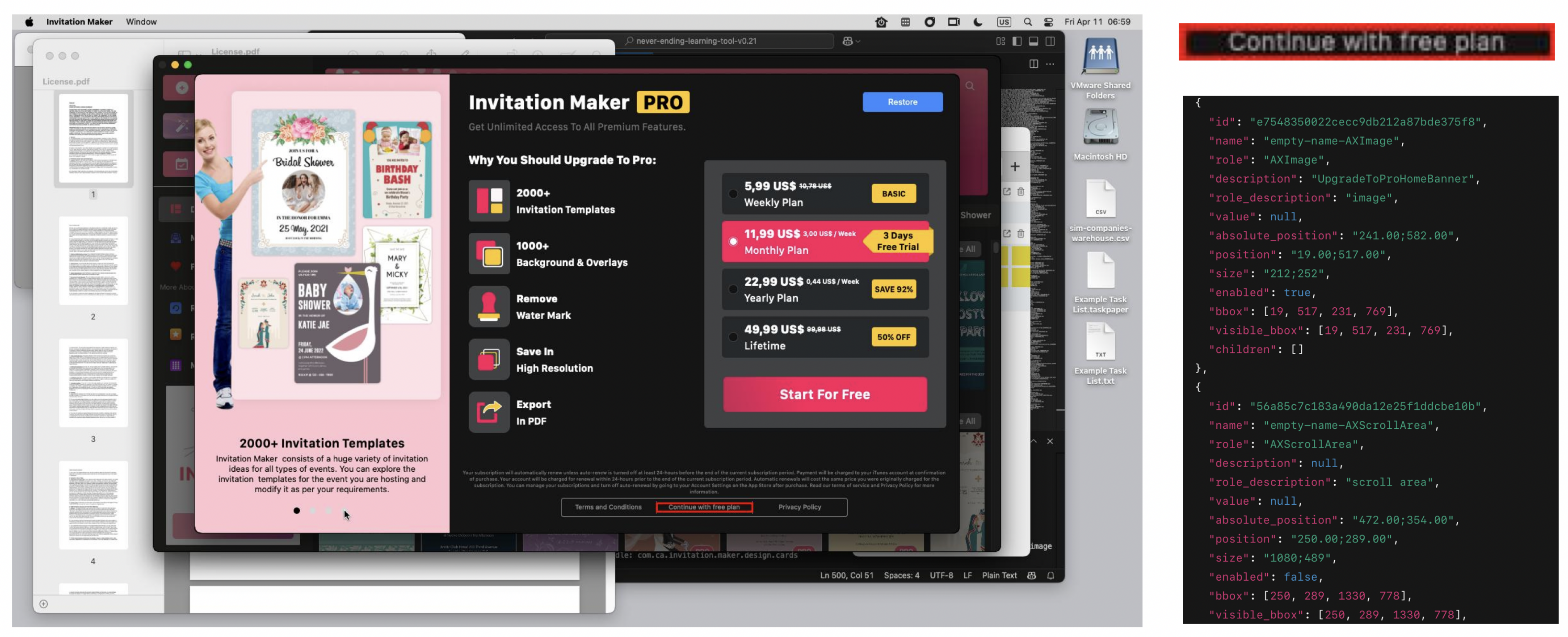}
    \caption{Example sample from our dataset. Left: screenshot of the macOS desktop interface. Upper right:  target element cropped. Lower right: a segment of the accessibility tree.
    }
    \label{fig:sample_from_the_dataset}
\end{figure}

\subsubsection{Collection statistics}

\begin{figure}[H]
    \centering
    \includegraphics[width=0.9\linewidth]{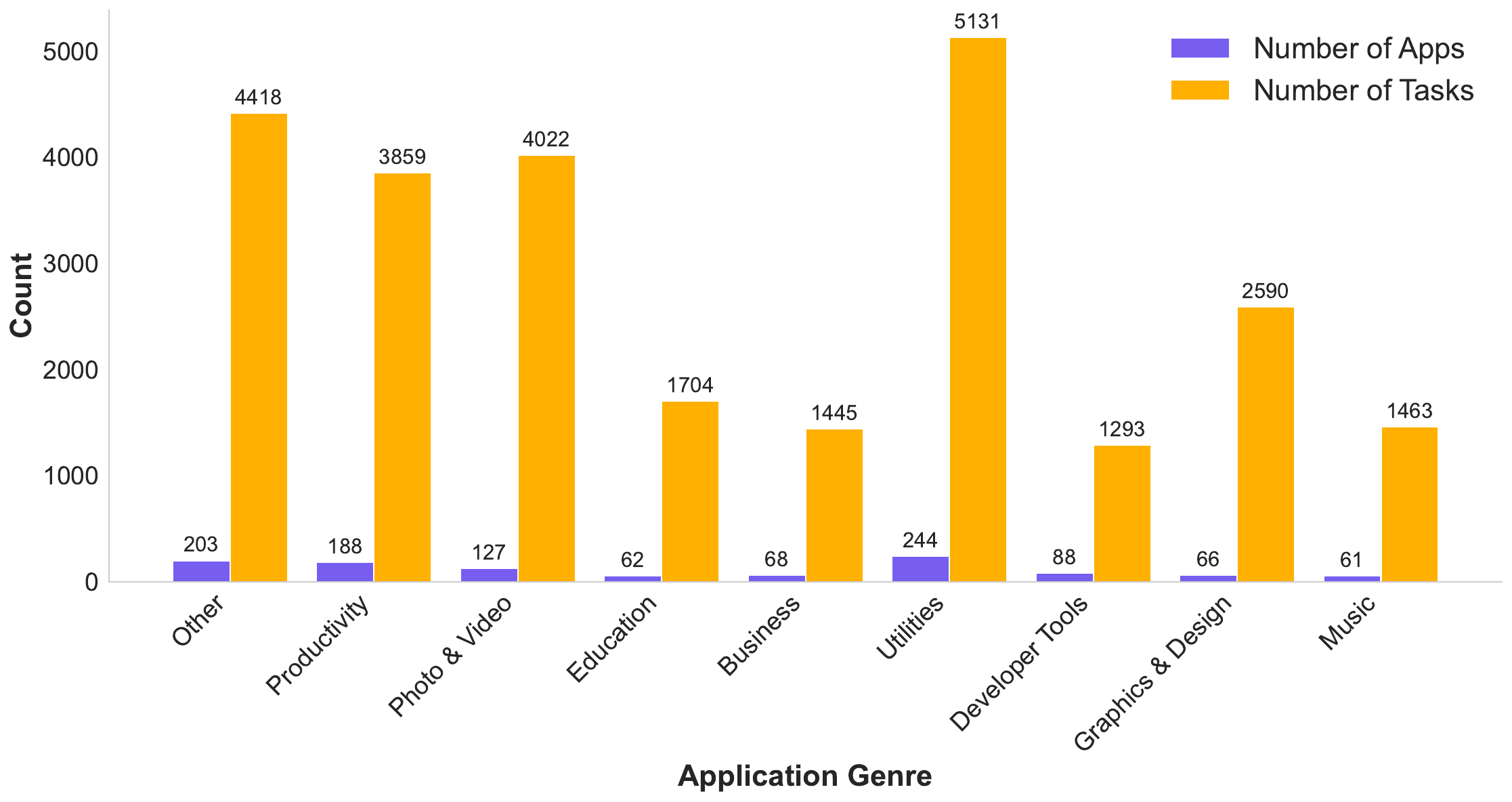}
    \caption{Number of apps and associated tasks per genre. For each genre, the left bar shows the number of apps, and the right bar shows the number of tasks. Colors distinguish between the two quantities. The figure highlights disparities between app availability and task density across categories.}
    \label{fig:applications-tasks-per-category}
\end{figure}

\begin{figure}[H]
    \centering
    \begin{subfigure}[t]{0.48\textwidth}
        \centering
        \includegraphics[width=\textwidth]{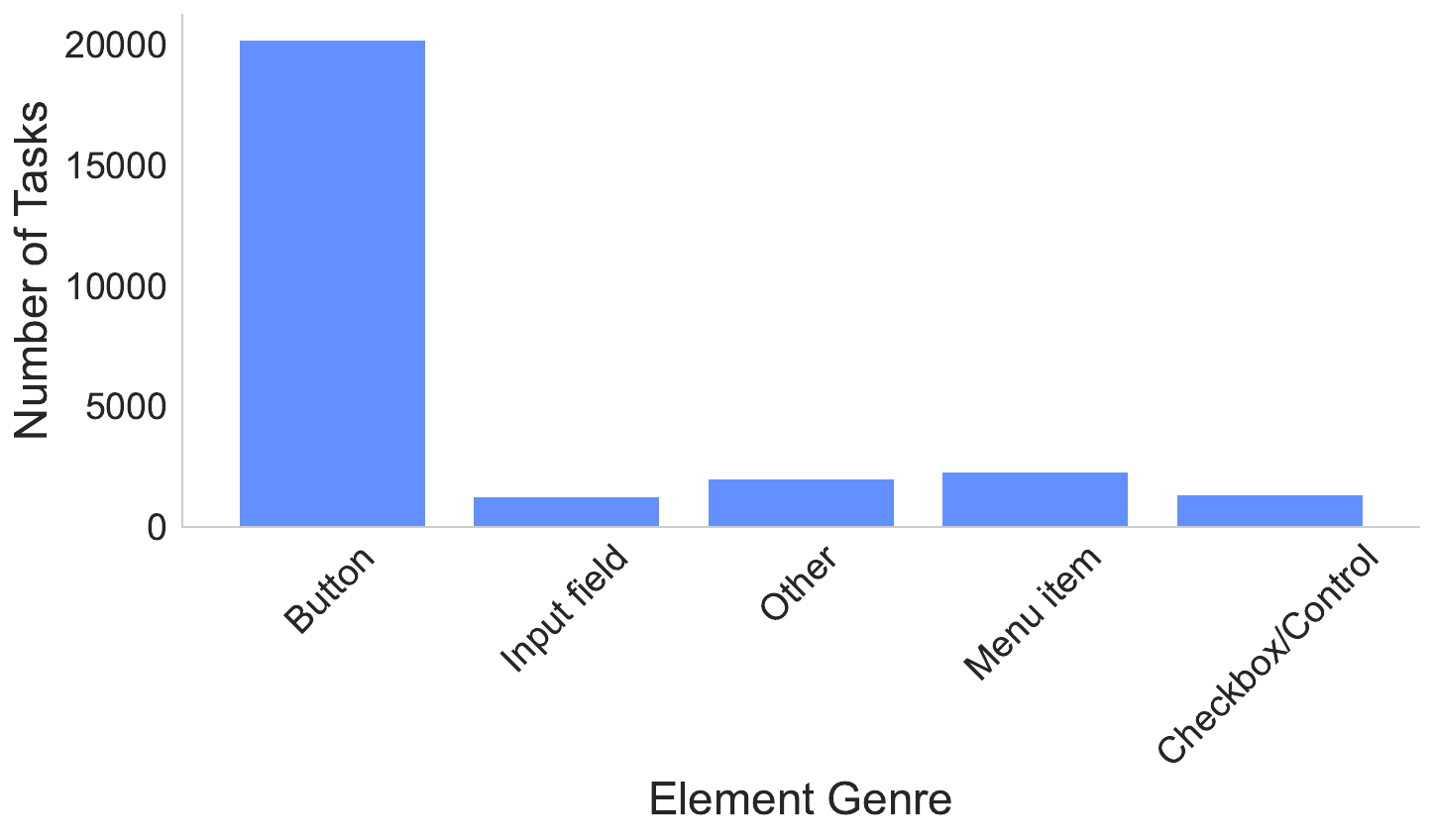}
        \caption{Distribution of tasks per element type. The most prevalent category is buttons.}
        \label{fig:tasks_per_element}
    \end{subfigure}
    \hfill
    \begin{subfigure}[t]{0.48\textwidth}
        \centering
        \includegraphics[width=\textwidth]{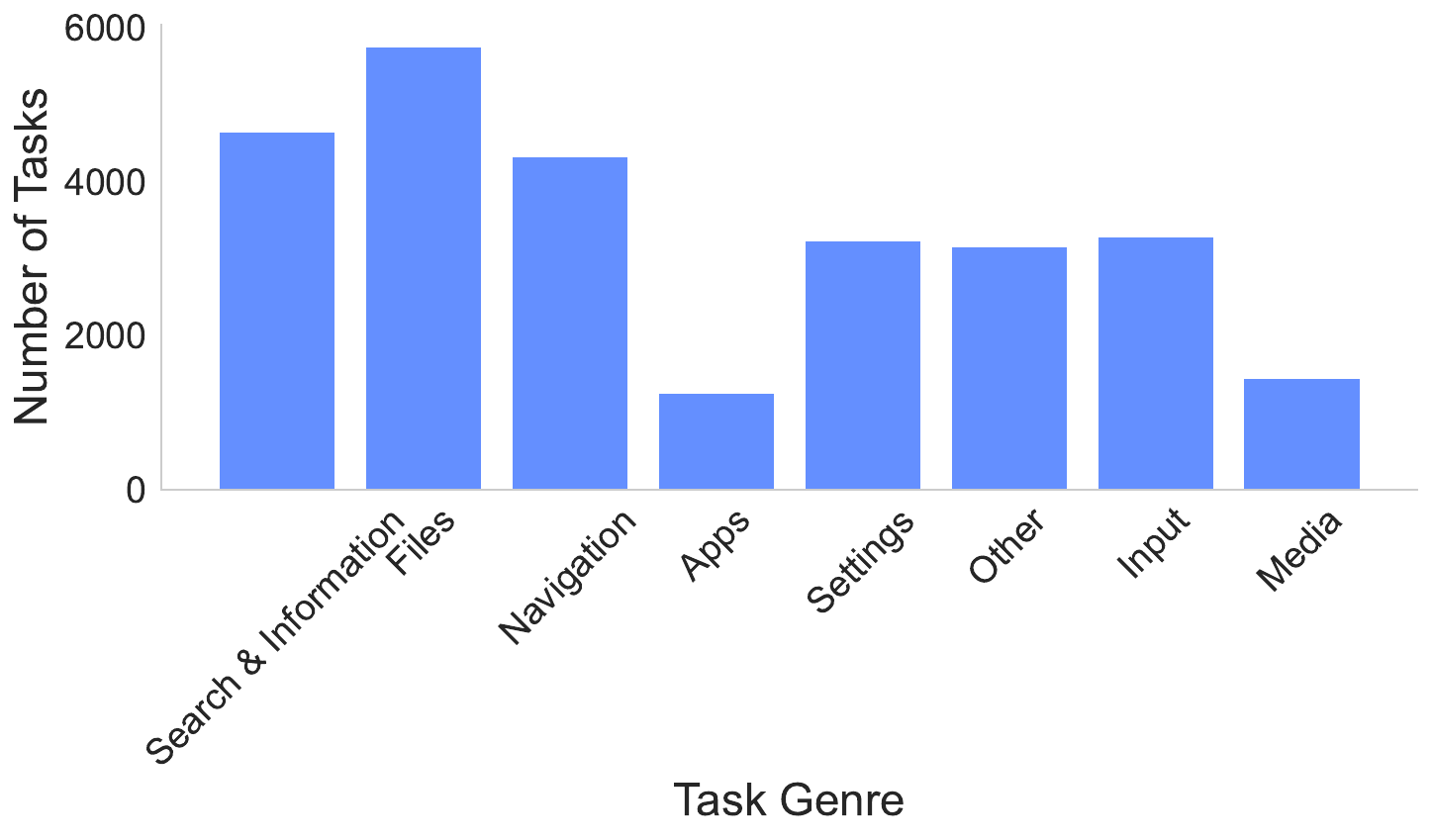}
        \caption{Distribution of tasks per task type.}
        \label{fig:tasks_per_type}
    \end{subfigure}
    \caption{Distributions of tasks across element types (left) and task types (right).}
    \label{fig:task_distributions}
\end{figure}

\begin{figure}[H]
    \includegraphics[width=1.2\linewidth]{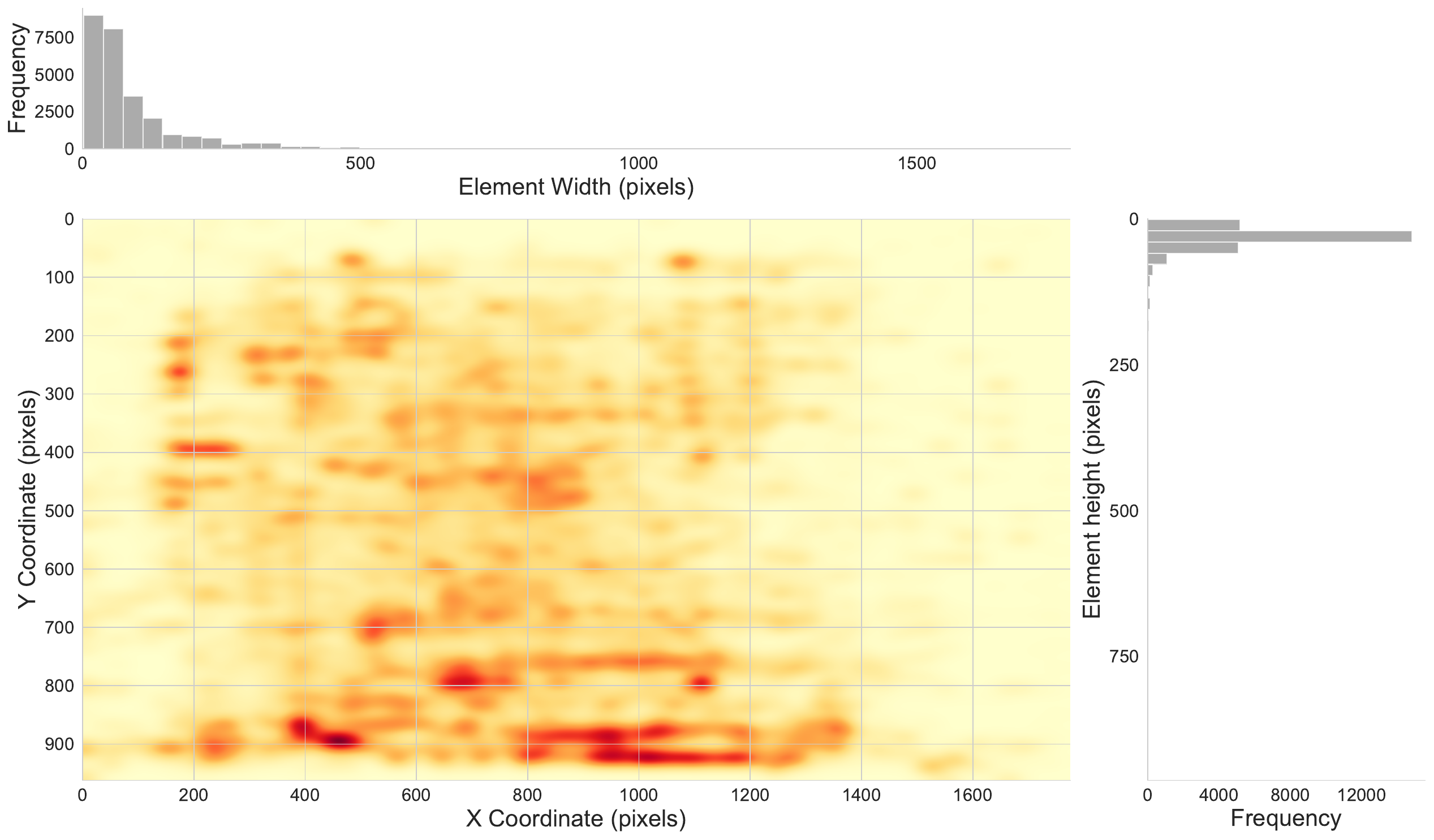}
    \caption{Top-left: Distribution of target element widths. Bottom-left: Distribution of target element center locations, showing that most target elements are positioned near the bottom of the desktop interface. Bottom-right: Distribution of target element heights.}
    \label{fig:heatmap}
\end{figure}

Many interaction targets are located in peripheral regions (e.g., toolbars, corners), and a large proportion are visually small, with limited surface area (Figure~\ref{fig:heatmap}).

\subsection{Parameters}
\label{appendix:parameters}

\begin{table}[H]
\centering
\begin{tabular}{lp{9cm}}
\toprule
\textbf{Parameter} & \textbf{Description} \\
\midrule
Maximum parsing duration & Specified in minutes, default is 2 hours \\
Deterministic text input & Default string is \texttt{'DEFAULT'} \\
Maximum parsing tree depth & Default is 25 \\
Cursor move before click & Defaults to \texttt{False} \\
Agent usage & Set to \texttt{True} by default. Can be enabled if an OpenAI API key is provided in a separate file \\
Task collection & Defaults to \texttt{True}. If set to \texttt{False}, trees can be collected without their associated tasks \\
\bottomrule
\end{tabular}
\vspace{0.1in}
\caption{Configuration Parameters for the Crawler}
\label{tab:crawler-config}
\end{table}

\subsection{Prompts}
\label{appendix:prompts}

\begin{tcolorbox}[title=Input Agent Instructions,colback=blue!5!white,colframe=blue!75!black,breakable]
Analyze the given macOS application accessibility screen information and follow these steps:

\begin{enumerate}
    \item Determine the type and purpose of the application based on the provided elements and descriptions.
    \item Identify all \texttt{AXTextField} elements present in the structure.
    \item For each \texttt{AXTextField}:
    \begin{enumerate}
        \item Infer its specific purpose within the application context.
        \item Consider what a user would input in this field based on accessibility cues and typical behavior.
        \item Generate an example input relevant to the field’s likely function and the app’s overall purpose.
    \end{enumerate}
\end{enumerate}

\textbf{Output:} A JSON object where:
\begin{itemize}
    \item \textbf{Keys} = integer IDs of the \texttt{AXTextField} elements
    \item \textbf{Values} = realistic example inputs, based on screen context
\end{itemize}

Only return the JSON object—no additional explanations.

\textbf{Examples:}
\begin{itemize}
    \item \texttt{\{7: "Yellow Submarine"\}} \hfill // Music app search
    \item \texttt{\{12: "John", 15: "Smith", 21: "07580198241"\}} \hfill // Contacts app
    \item \texttt{\{8: "main"\}} \hfill // IDE project file search
\end{itemize}

\textbf{Note:} Ensure that inputs are app-appropriate and reflect common human interactions.
\end{tcolorbox}

\begin{tcolorbox}[title= Order Agent Instructions,colback=yellow!5!white,colframe=yellow!80!black,breakable]
Given accessibility screen info, organize UI elements in logical interaction order. Consider irreversible actions and screen transitions.

\textbf{Output:} JSON with nested groups (max 8), each containing element IDs:
\begin{itemize}
    \item Prioritize elements in popovers, content switches, and window controls.
    \item Derive element type from description if needed.
    \item Include \textbf{ALL element IDs} from input.
\end{itemize}

\textbf{Grouping Rules:}
\begin{itemize}
    \item \texttt{dynamic\_TYPE} — dynamic lists (emails, notes, etc.)
    \item \texttt{repeated\_TYPE} — options where only one is needed (date, category, etc.)
    \item Avoid grouping unrelated or static UI items together.
\end{itemize}

\textbf{Flags to include when relevant:}
\begin{itemize}
    \item \texttt{"login\_page": true}
    \item \texttt{"system\_access\_required": true}
\end{itemize}

\vspace{2mm}
\textbf{Example 1 — Complex App:}
\begin{verbatim}
{
  "action_order": [
    {"menu_buttons": [1, 2, 3]},
    {"dynamic_emails": [4, 5, 6, 7]},
    {"repeated_time_selection": [8, ..., 31]},
    {"popover_buttons": [32, 33]}
  ],
  "login_page": false,
  "system_access_required": false
}
\end{verbatim}

\textbf{Example 2 — Login Page:}
\begin{verbatim}
{
  "action_order": [
    {"login_elements": [1, 2, 3]},
    {"account_settings": [4]}
  ],
  "login_page": true,
  "system_access_required": false
}
\end{verbatim}
\end{tcolorbox}

\begin{tcolorbox}[title=Click Task Prompt, colback=purple!5!white, colframe=purple!60!black, fonttitle=\bfseries, boxrule=0.8pt,breakable]
\small
You are given a UI screenshot, an image of the clicked UI element. 
The clicked element is highlighted in red.
Your task is to describe the action needed to click this element.

Guidelines:

    0. If the element is not perfectly selected (ex. partially), the box is strangely located, or no human would do this task - return empty string.

    1. The task must describe the function, not the appearance of the element.
    For example, prefer "Create a new document" over "Click the grey + button." Repeating the element's text is acceptable.

    2. The task must be unique to this screen.
    For example, if there are two buttons labeled "Open," you must specify which "Open" button is meant.
    
    3. The task must consider the app context, but not imagine extra information.
    For example, if the app is an image editor and the button is "Delete," the better task is "Delete an image", not just a generic "delete."

    4. Use the fewest words possible without sacrificing clarity.

    5. Write the task in straightforward English only.

    6. Select a category for each task. Must be one of Navigation (go back), Settings (adjust volume), Files (save file), Apps (open edge), Search \& Information (check weather), Media (play music), Accounts (sign in), Communication (share file), Input (enlarge font), Connectivity (connect wifi), Modes (dark mode), E-commerce (add to cart)

    7. Select a category for each element. Must be one of Image, Text, Checkbox/Control, Menu item, Input field, Button, Group, Link.

Important notes:

    The click is based on accessibility information. Metadata may be incorrect or the element may not exist.

    Rely primarily on the images.

    The element image should show a single element with a unique function.
    If the element is obstructed, covered by a window or pop-up, or if multiple cropped elements are shown — return an empty string.

    Inspect the red box carefully: if the element is not visible, return an empty string.

    If there is no red box - return empty string.
Return your answer in JSON format, with no extra text.

Example:
\begin{verbatim}{
    "task": "Open the menu to see tutorials", 
    "task_category": "Search & Information", 
    "element_category": "Button"}
\end{verbatim}
\end{tcolorbox}

\begin{tcolorbox}[title=Input Task Prompt, colback=orange!5!white, colframe=orange!60!black, fonttitle=\bfseries, boxrule=0.8pt,breakable]
\small
\textbf{You are given:}
\begin{itemize}
  \item An original task description for a UI interaction: \texttt{\{task\_string\}}
  \item A screenshot showing the full interface with a red-highlighted element
  \item A cropped view focusing on just the highlighted element
\end{itemize}

\textbf{Your goal:} Change the task into a natural-language instruction \textbf{fully in English} that involves only inputting text. Output an action as \texttt{"type"} + the exact text to input. If not clearly solvable from the task, revise it.

\textbf{Key Principles:}
\begin{itemize}
  \item Make it sound like a real instruction a person would give
  \item Use exact input (no placeholders); don’t interpret content—be explicit
  \item Focus on real-world intent and what a user is likely trying to do
\end{itemize}

\textbf{Requirements:}
\begin{itemize}
  \item Instruction must be clear, natural, and concise
  \item Action must start with \texttt{type} and include exact text
  \item Both fields must be fully in English
  \item No placeholders like “your name” or “email”
  \item Avoid click/press/select – only typing
  \item Must be obvious what to type from the instruction
  \item Never add phrases like “by typing it”
\end{itemize}

\textbf{Output Format (JSON):}
\begin{verbatim}
{"task": "Use john.doe@example.com as your login email",
 "action": "type john.doe@example.com"}
\end{verbatim}

\textbf{Bad vs Good Examples:}
\begin{itemize}
  \item “Enter coded message” → “Enter 1234 as your coded message”
  \item “Save your converted files…” → “Use /Users/yourname/Desktop as your destination folder”
  \item “Check the box labeled…” → “Select Include borders and shadings as your option”
\end{itemize}

\textbf{Avoid These Mistakes:}
\begin{itemize}
  \item Placeholder text: “your name” → “Maria Garcia”
  \item Mechanical: “password in field” → “Use TrustNo1 as your password”
  \item UI-only focus: “Fill search box” → “Find information about electric cars”
  \item Vague: “Type the code” → “Enter 8294 as your verification code”
  \item Impersonal: “Input required” → “Add your birthday as 03/15/1988”
\end{itemize}
\end{tcolorbox}

\subsection{Training Setup}
\label{appendix:training}

We training 3 GUI agents on the collected dataset.

\subsubsection{GUIrilla-See-0.7B}

\emph{GUIrilla-See-0.7B} is built on \textsc{Florence 2-Large} ($\approx$ 0.7 B parameters) and fine-tuned via supervised fine-tuning for \emph{open-vocabulary detection} in GUI screenshots.  
Given an image and a free-form textual query, the model predicts either a bounding box or a polygon mask that encloses the best-matching UI element.

\paragraph{LoRA configuration.}
Fine-tuning uses Low-Rank Adaptation with RSLoRA initialisation:

\begin{itemize}
    \item rank $r = 8$
    \item scaling $\alpha = 16$
    \item dropout $= 0.05$
    \item bias $= \textit{none}$
    \item target modules $= \{\texttt{q\_proj}, \texttt{o\_proj}, \texttt{k\_proj}, \texttt{v\_proj}, \texttt{linear}, \texttt{Conv2d}, \texttt{lm\_head}, \texttt{fc2}\}$
    \item weight init \textit{Gaussian}
\end{itemize}

\paragraph{Training setup.}
\begin{itemize}
    \item Hardware: 1 × NVIDIA A100 40 GB.
    \item Batch size: 8,  mixed precision.
    \item Optimiser: AdamW, learning rate $2\times10^{-5}$. Cosine decay schedule with a 5\% warm-up fraction.
    \item Epochs: 4; total wall-clock time $\approx$ 10 hours.
\end{itemize}

\subsubsection{GUIrilla-See-3B}

\emph{GUIrilla-See-3B} starts from \textsc{Qwen-2.5-VL-3B-Instruct} (3 B parameters) and is fine-tuned with supervised fine-tuning (SFT) to localise macOS GUI elements.  
Given a full-desktop screenshot and a natural-language instruction, the model outputs a single coordinate $(x,y)$ that lies at (or very close to) the centre of the referenced region.

\paragraph{LoRA configuration.}
Fine-tuning uses Low-Rank Adaptation (LoRA) in \texttt{bfloat16} mixed precision:

\begin{itemize}
    \item rank $r = 32$
    \item scaling $\alpha = 16$
    \item dropout $= 0.1$
    \item bias $= \textit{none}$
    \item target modules $= \{\texttt{down\_proj},\texttt{o\_proj},\texttt{k\_proj},\texttt{q\_proj},\texttt{gate\_proj},\texttt{up\_proj},\texttt{v\_proj}\}$
    \item weight init \textit{Gaussian}
\end{itemize}

\paragraph{Training setup.}
\begin{itemize}
    \item Hardware: \textbf{2 $\times$ NVIDIA H100 80 GB}.
    \item Global batch size: 16
    \item Optimiser: AdamW with $\beta_{1}=0.9,\;\beta_{2}=0.95$.
    \item Learning rate: $2\times10^{-5}$, cosine decay schedule, warm-up ratio $0.05$.
    \item Attention kernel: \textbf{FlashAttention-2} for memory-efficient training.
    \item Epochs: 2; total wall-clock time $\approx$ 5 hours.
\end{itemize}

\subsubsection{Fine-tuning improvement on base models.}

\begin{table*}[htbp]
\centering
\scriptsize
\renewcommand{\arraystretch}{1.1}
\begin{tabular}{lcc}
\toprule
\textbf{Model} & \textbf{Base Acc. (\%)} & \textbf{Tuned Acc (\%)} \\ 
\midrule
Florence Large (0.7B) & 8.31 & 53.55 \\
Qwen 2.5 VL (3B) & 18.40 & 73.48 \\
Qwen 2.5 VL (7B) & 35.78 & \textbf{75.59} \\
\bottomrule
\end{tabular}
\vspace{0.05in}
\caption{Accuracy improvements after fine-tuning on \textsc{GUIrilla-Task}.}
\label{tab:tuned-acc}
\end{table*}

\subsubsection{GUIrilla-See-7B}

We also train a larger model that starts from \textsc{Qwen-2.5-VL-7B-Instruct} (7B parameters).  
All LoRA, optimiser, and scheduler settings are kept \emph{identical} to the 3B run.  
Using the same 2 $\times$ H100 80 GB configuration with FlashAttention-2, training finishes in roughly 6–7 hours.

\subsection{ScreenSpot Details} 

\subsubsection{Data Leakage Analysis on ScreenSpot} \label{appendix:data-leakage}
We manually screened overlaps by bundle IDs and application names to ensure no data leakage happened during training for both ScreenSpot-v2 and ScreenSpot-Pro benchmarks. As ScreenSpot-v2 doesn’t provide this information, we manually labeled the apps there.  

We discovered the following overlaps out of 881 applications in our train dataset and ScreenSpot test sets:
OneNote appears in train data (macOS app) and in ScreenSpot-v2 (Windows). This app has 1 task in the benchmark, and the login screen looks identical, so the leakage may have affected the result. We adjusted the score to account for it from 90.41\% -> 90.33\%. This doesn’t influence the ranking, yet we adjusted the score for fairness.
Microsoft Excel appears in the train dataset and in ScreenSpot-Pro. Here we manually looked at every screen (screen ids 4650-4659) and found that our data only includes a login flow and never actually opens the main app and its functionality. In ScreenSpot-Pro on the other hand, all tasks focus on Excel functions as part of the multi screen window. So, we assume that no major leakage was done here.

\begin{figure}[H] 
  \centering
  \begin{subfigure}{0.49\linewidth}
    \includegraphics[width=\linewidth]{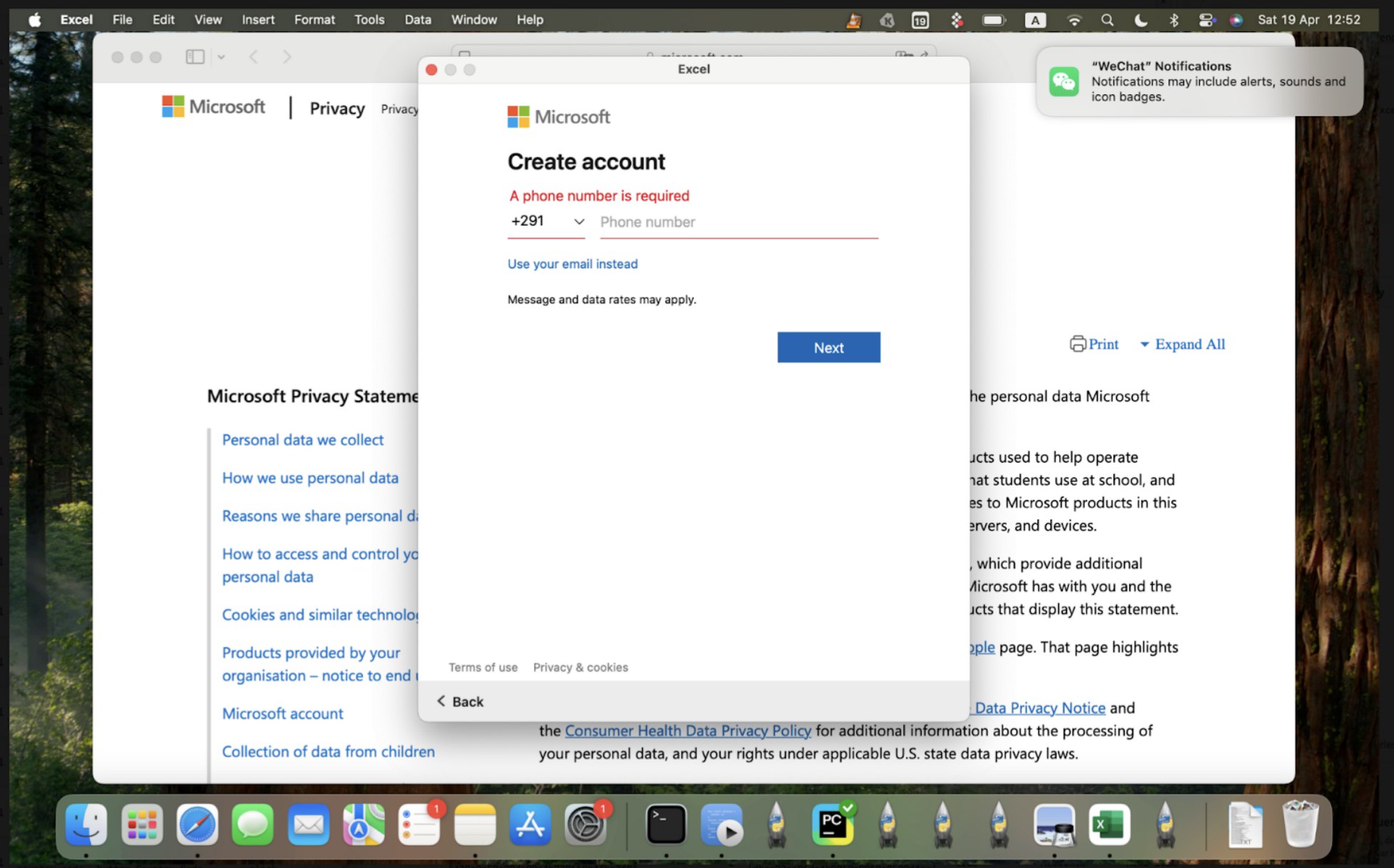}
    \caption{GUIrilla: Excel login page.}
    \label{fig:excel-guirilla}
  \end{subfigure}
  \hfill
  \begin{subfigure}{0.49\linewidth}
    \includegraphics[width=\linewidth]{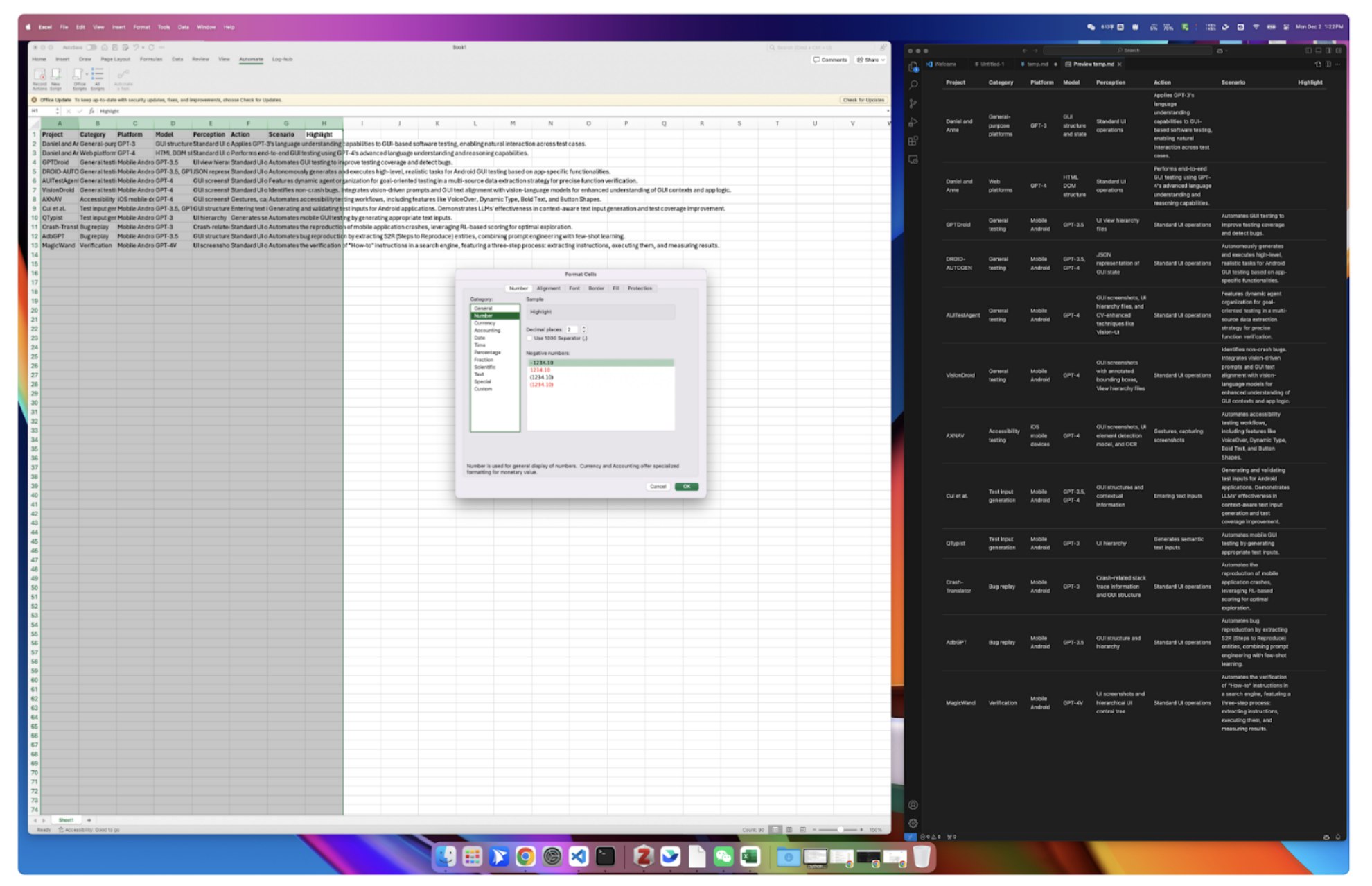}
    \caption{ScreenSpot-Pro: Main app with table manipulation.}
    \label{fig:excel-screenspot}
  \end{subfigure}
  \caption{Side-by-side comparison of Excel app data across datasets.}
  \label{fig:excel-side-by-side}
\end{figure}

\begin{table*}[!ht]
\centering
\resizebox{\textwidth}{!}{
\begin{tabular}{l c c c c c c}
\toprule
\textbf{Model} & \textbf{Development} & \textbf{Creative} & \textbf{CAD} & \textbf{Scientific} & \textbf{Office} & \textbf{Overall Acc} \\
\midrule
UI-TARS-72B       & 40.8  & 39.6  & 17.2  & 45.7  & 54.8  & 38.1 \\
UI-TARS-7B        & 36.1  & 32.8  & 18.0  & 50.0  & 53.5  & 35.7 \\
GUIrilla-See-7B   & 30.10 & 31.96 & 26.82 & 47.64 & 53.04 & 35.36 \\
GUIrilla-See-3B   & 24.08 & 25.51 & 24.52 & 39.37 & 47.39 & 29.35 \\
UI-TARS-2B        & 26.4  & 27.6  & 14.6  & 39.8  & 42.6  & 27.7 \\
Qwen 2.5 VL (7B)  & 24.7  & 24.9  & 14.6  & 30.3  & 47.0  & 27.1 \\
OS-Atlas-7B       & 17.7  & 17.9  & 10.3  & 24.4  & 27.4  & 18.9 \\
UGround-7B        & 14.7  & 17.0  & 11.1  & 19.3  & 27.0  & 16.5 \\
Qwen 2.5 VL (3B)  & 10.7  & 17.3  & 7.3   & 24.4  & 28.7  & 15.9 \\
CogAgent (18B)    & 8.0   & 5.6   & 6.1   & 13.4  & 10.0  & 7.7 \\
ShowUI (2B)       & 9.4   & 5.3   & 1.9   & 10.6  & 13.5  & 7.7 \\
GUIrilla-See-0.7B & 6.69  & 6.45  & 3.83  & 12.20 & 8.26  & 7.34 \\
OS-Atlas-4B       & 3.7   & 2.3   & 1.5   & 7.5   & 4.8   & 3.7 \\
Florence-2 (0.7B) & 0.0   & 0.0   & 0.0   & 0.4   & 0.0   & 0.12 \\
\bottomrule
\end{tabular}}
\caption{Task category performance per app category on ScreenSpot-Pro.}
\label{app-category}
\end{table*}

\newpage
\subsection{Additional Evaluation Details}

\subsubsection{Grounding}
\label{sec:evaluation}

A detailed quantitative comparison with existing datasets is provided in Table~\ref{tab:dataset-comparison}.

\begin{table}[htbp]
\centering
\resizebox{\textwidth}{!}{%
\begin{tabular}{l@{\hspace{2pt}}c@{\hspace{2pt}}c@{\hspace{2pt}}c@{\hspace{3pt}}l@{\hspace{3pt}}c@{\hspace{3pt}}c@{\hspace{3pt}}c@{\hspace{3pt}}c@{\hspace{3pt}}c}
\toprule
\textbf{Dataset} & \textbf{\#Apps} & \textbf{\#Tasks} & \textbf{\#Unique UIs} & \textbf{Collection} & \textbf{Desktop} & \textbf{macOS} & \textbf{Fullscreen} & \textbf{Grounding} & \textbf{Agentic}\\
\midrule
OSWorld~\citep{xie2024osworldbenchmarkingmultimodalagents} & 10 & 369 & - & Manual & $\checkmark$ & $\times$ & $\checkmark$ & $\times$ & $\checkmark$ \\
OmniACT~\citep{Kapoor2024OmniACT} & 60 & 9802 & - &  Manual & $\checkmark$ & $\checkmark$ & $\checkmark$ & $\times$ & $\times$ \\
ScreenSpot-V2~\citep{os-atlas} & 6 & 324 & 187  & Manual & $\checkmark$ & $\times$ & $\times$ & $\checkmark$ & $\times$ \\
ScreenSpot-Pro~\citep{Li2025ScreenSpot} & 23 & 1581 (511) & -  & Manual & $\checkmark$ & $\checkmark$ & $\checkmark$ & $\checkmark$ & $\times$ \\
Mind2Web~\citep{Deng2023Mind2Web} & $\times$ & 2350 & 2350  & Manual & $\times$ & $\times$ & $\times$ & $\checkmark$ & $\checkmark$ \\
OSAtlas~\citep{os-atlas} & - & - & 2.2M (1339) & Automated & $\checkmark$ & $\checkmark$ & $\checkmark$ & $\checkmark$ & $\times$ \\
Web-Hybrid~\citep{UGround-Web-Hybrid} & $\times$ & - & 773K  & Automated  & $\times$ & $\times$ & $\times$ & $\checkmark$ & $\times$ \\
\textbf{GUIrilla-Task} & \textbf{1108} & \textbf{27171} & \textbf{4179}  & Automated & $\checkmark$ & $\checkmark$ & $\checkmark$ & $\checkmark$ & $\checkmark$ \\
\bottomrule
\end{tabular}%
}
\begin{tablenotes}
\small
\item The number in brackets denotes the reported quantity for macOS.
\end{tablenotes}
\vspace{0.1in}
\caption{Comparison of Existing Datasets for Task Automation}
\label{tab:dataset-comparison}
\end{table}

We fine-tune and release three \textsc{GUIrilla-See} agents of varying parameter scales on our \textsc{GUIrilla-Task} dataset: \textsc{GUIrilla-See} (0.7B) (based on Florence-2-large \citep{xiao2024florence}), \textsc{GUIrilla-See} (3B) and \textsc{GUIrilla-Se}e (7B) (based on Qwen-2.5-VL-Instruct \citep{bai2025qwen2}). All models are trained exclusively on \textsc{GUIrilla-Task} dataset. For training details, see Appendix \ref{appendix:training}. 

\textbf{macOS Grounding Evaluation}. We evaluate element localization by functional category on the GUIrilla-Task test set and compare against representative multi-OS baselines (UI-TARS, OS-Atlas, UGround). Rather than positioning this as a race for state-of-the-art, we use these results to illustrate how \textsc{GUIrilla-Task} provides practical, function-level supervision for realistic macOS desktop scenarios. We find particularly strong gains in Settings (+8.7), Connectivity (+26.3), Files (+7.5), and Input (+8.7), as shown in Table \ref{tab:task-category-performance-transposed-nodevices}.
Importantly, improvements are spread across element types as well (buttons, input fields, dialogs), the full table can be found in Appendix, Table \ref{tab:grounding}.

\begin{table*}[h]
\centering
\resizebox{\textwidth}{!}{
\begin{tabular}{l *{12}{c}}
\toprule
\textbf{Model} & \textbf{Communication} & \textbf{Files} & \textbf{Navigation} & \textbf{Search \& Information} & \textbf{E-commerce} & \textbf{Accounts} & \textbf{Input} & \textbf{Apps} & \textbf{Media} & \textbf{Settings} & \textbf{Connectivity} & \textbf{Total} \\
\midrule
UI-TARS 2B   & 27.6\% & 45.6\% & 53.3\% & 49.5\% & 52.2\% & 61.9\% & 31.3\% & 50.0\% & 35.3\% & 50.3\% & 42.1\% & 47.53\% \\
UI-TARS 1.5 7B   & 48.3\% & 67.0\% & 63.9\% & 74.7\% & 72.6\% & \textbf{81.0\%} & 56.5\% & 68.8\% & 54.9\% & 80.9\% & 68.4\% & 69.07\%\\
OS-Atlas 7B       & 48.3\% & 64.9\% & 59.9\% & 68.3\% & 70.8\% & \textbf{81.0\%} & 53.9\% & 66.7\% & \textbf{60.8\%} & 66.5\% & 63.2\% & 64.86\% \\
UGround 2B   & 51.7\% & 63.0\% & 60.8\% & 70.0\% & 69.0\% & \textbf{81.0\%} & 45.2\% & 62.5\% & 56.9\% & 67.6\% & 68.4\% & 64.03\% \\
UGround 7B   & 62.1\% & 67.4\% & 68.7\% & 75.4\% & 69.0\% & \textbf{81.0\%} & 54.8\% & 70.8\% & 52.9\% & 78.6\% & 52.6\% & 69.46\% \\
GUIrilla-See 3B  & 51.7\% & 74.7\% & 68.7\% & 77.8\% & 76.1\% & \textbf{81.0\%} & 57.4\% & \textbf{72.9\%} & \textbf{60.8\%} & 82.7\% & 73.7\% & 73.48\% \\
GUIrilla-See 7B  & \textbf{65.5\%} & \textbf{74.9\%} & \textbf{70.5\%} & \textbf{79.2\%} & \textbf{78.8\%} & \textbf{81.0\%} & \textbf{65.2\%} & 70.8\% & \textbf{60.8\%} & \textbf{87.3\%} & \textbf{78.9\%} & \textbf{75.59\%} \\
\bottomrule
\end{tabular}
}
\caption{Performance breakdown across task categories on GUIrilla-Task test set. Best performance per category shown in \textbf{bold}.}
\label{tab:task-category-performance-transposed-nodevices}
\end{table*}

\textbf{ScreenSpot Evaluation}. Table~\ref{tab:sspro_results} reports grounding accuracy on ScreenSpot-v2 \citep{Li2025ScreenSpot} and ScreenSpot-Pro. While absolute comparisons are influenced by differences in architectures, training pipelines, and closed-source data, the results provide context on how \emph{dataset composition and platform realism} affect transfer. Notably, \textsc{GUIrilla-See} (7B), trained on only 4.2K macOS full-desktop screens, achieves comparable macOS-specific accuracy to much larger multi-OS training regimes, despite using orders of magnitude fewer images. This supports the central point of the paper: \textsc{GUIrilla-Task} is a compact but high-utility macOS dataset that can be reused to train and stabilize downstream models. To ensure fairness in evaluation, we made sure that there is no data leakage, details can be found in Appendix \ref{appendix:data-leakage}. 

\begin{table*}[h]
\centering
\resizebox{\textwidth}{!}{
\begin{tabular}{l l l l l c c}
\toprule
\textbf{Model} & \textbf{Platform} & \textbf{Data} & \textbf{\# Images} & \textbf{ScreenSpot-v2} & \textbf{ScreenSpot-Pro} & \textbf{ScreenSpot-Pro \newline (macOS)}\\
\midrule

Florence-2 (0.7B) & Multi-OS & Real + Synthetic & - & 1.8\% & 0.12\% & 0.16\% \\

Qwen 2.5 VL (3B) & Multi-OS & Real + Synthetic & - & 62.34\% & 15.93\% & 18.37\% \\

Qwen 2.5 VL (7B)  & Multi-OS & Real + Synthetic & - & 87.5\% & 27.13\% & 34.27\% \\

\midrule

CogAgent (18B) & Multi-OS & Real + Synthetic & 40M & 52.8\% &7.7\% & 4.6\% \\

UGround (7B) & Web + Android & Real & 1.3M & 76.3\% & 16.5\% & 12.3\% \\

ShowUI (2B) & Multi-OS & Real + Synthetic & 256K & 77.3\% & 7.7\% & 10.8\% \\

OS-Atlas (7B) & Multi-OS & Synthetic & 2.2M & 83.3\% & 18.9\% & 20.0\% \\

UI-TARS (2B) & Multi-OS & Real + Synthetic & $\sim$20M (est.) & 84.7\% & 27.7\% & 15.4\% \\

UI-TARS (72B) & Multi-OS & Real + Synthetic & $\sim$20M (est.) & 90.3\% & \textbf{38.1\%} & \underline{40.0\%} \\

UI-TARS (7B) & Multi-OS & Real + Synthetic & $\sim$20M (est.) & \underline{91.6\%} & \underline{35.7\%} & 27.7\% \\

\midrule

GUIrilla-See (0.7B) & macOS & Synthetic & \textbf{4.2K} & 53.55\% & 7.34\% & 7.95\% \\

GUIrilla-See (3B) & macOS & Synthetic & \textbf{4.2K}  & 89.54\% & 29.35\% & 32.62\% \\

GUIrilla-See (7B) & macOS & Synthetic & \textbf{4.2K} & \textbf{94.73\%} & 35.36\% & \textbf{41.39\%} \\

\bottomrule
\end{tabular}
}
\vspace{0.1in}
\caption{Grounding accuracy comparison on ScreenSpot-v2 and ScreenSpot-Pro (full and macOS subset). \textbf{Bold} is used to highlight the best result, \underline{underline} to hightlight second best result.}
\label{tab:sspro_results}
\end{table*}

Additionally, we see that full-screen supervision (compared to \citep{UGround-Web-Hybrid}) as well as task formulation on a function level (compared to description-only as in \citep{os-atlas}) can enable better contextual grounding in realistic GUI settings. Additional analysis of model robustness across different decoding strategies and confidence intervals are provided in Appendix~\ref{app:confidence-intervals}.

In grounding evaluations (Table~\ref{tab:grounding}), \textsc{GUIrilla-See} (7B) achieved the highest overall accuracy at 75.59\%. \textsc{GUIrilla-See} (7B) showed particularly strong results on buttons, input fields, and "Other" elements such as icons and links, demonstrating robust performance across varied UI components. Interestingly, UGround shows marginal advantages in menu-heavy tasks, and we found their data to contain 400× more menu samples, reflecting the limits of scale without functional task diversity.

\begin{table*}[h]
\centering
\scriptsize
\renewcommand{\arraystretch}{1.1}
\begin{tabular}{lcccccc} 
\toprule
 & \multicolumn{6}{c}{\textbf{Grounding Accuracy (\%)}} \\ 
\textbf{Model} & Button & Input & Menu & Checkbox & Other & Overall \\  
\midrule
Qwen 2.5 3B & 8.0 & 0.0 & 0.0 & - & -- & -\\ 
Qwen 2.5 7B & 36.49 & 30.36 & 46.0 & 44.68 & 24.74 & 35.78\\ 
UI TARS 2B & 50.56 & 25.0 & 52.67 & 59.57 & 36.84 & 47.53\\
UGround v1 2B & 64.26 & 48.21 & 79.33 & 68.09 & 58.95 & 64.03\\ 
OS-Atlas-Base-7B & 65.76 & 53.57 & 72.67 & 57.45 & 62.11 & 64.86\\ 
UI TARS 1.5 7B & 68.57 & 64.29 & 86.67 & \textbf{78.72} & 58.42 & 69.07\\ 
UGround v1 7B & 68.67 & 56.25 & \textbf{88.67} & \textbf{78.72} & 64.21 & 69.46\\ 
\textbf{GUIrilla-See (3B)} & 74.48 & 61.61 & 81.33 & \textbf{78.72} & \textbf{66.84} & 73.48\\
\textbf{GUIrilla-See (7B)} & \textbf{76.55} & \textbf{66.07} & 86.67 & 76.6 & \textbf{66.84} & \textbf{75.59}\\ 
\bottomrule
\end{tabular}
\vspace{0.05in}
\caption{Grounding accuracy across models and element categories on \textsc{GUIrilla-Task}.}
\label{tab:grounding}
\end{table*}

\begin{figure*}[!ht]
    \centering
    \includegraphics[width=0.7\textwidth]{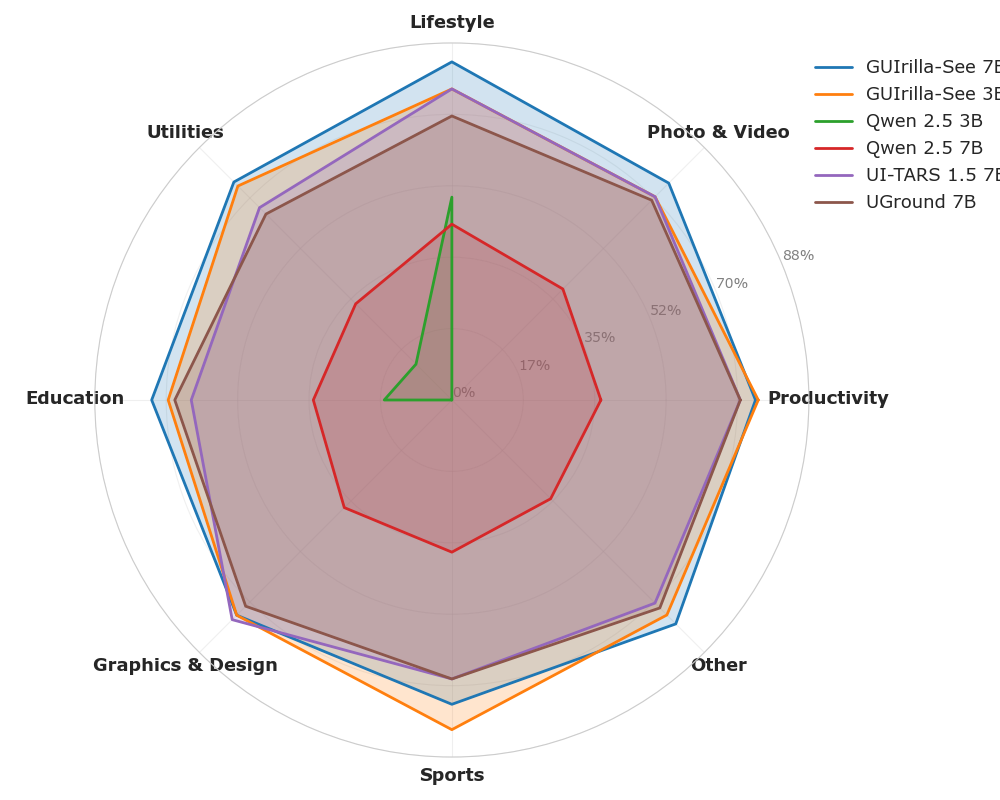}
    \caption{Comparative Performance on GUIrilla-Task (Grounding) of Vision Language Models Across Application Domains.  Larger models (7B) generally outperform smaller ones, with the biggest gains in Developer Tools, Productivity, and Graphics \& Design. GUIrilla-See 3B shows strong performance relative to 7B models, indicating effective domain specialization. }
\end{figure*}

\textbf{Cross-OS transfer.} Despite macOS-only training, \textsc{GUIrilla-See} (7B) reaches 32.03\% on Windows (ScreenSpot-Pro) and 41.39\% on macOS, exceeding OS-Atlas 7B (12.3\% / 20\%) and UGround 7B (14.9\% Windows). Thus, single-platform, function-level supervision does not preclude transfer and can outperform larger mixed-OS synthetic sets on challenging full-desktop scenes.

\subsubsection{Qualitative Analysis}
Analysis of 1,565 tasks across 227 applications reveals that macOS-specific training yields consistent improvements across fundamental UI interaction patterns. We identified 79 tasks where GUIrilla succeeds while all baselines fail, demonstrating strong understanding of macOS-specific paradigms: Finder-style dialogs ("Browse for movie destination folder"), System Preferences ("Edit advanced output settings"), and window management ("Close the Chat-with-Erix panel"). These success patterns validate our function-oriented approach, showing models learn what UI elements do rather than where they appear or their visual description.

\textbf{Failure Mode Analysis.} 
ScreenSpot-Pro evaluation reveals two key weaknesses: (1) icon-dense engineering tools such as Vivado, where tasks like “click group by repository button” or “open TCL console” fail due to limited representation of compact, icon-heavy UIs in the dataset; and (2) creative software like Illustrator and DaVinci Resolve, where canvas-focused actions such as “draw a circle” or “select brush tool” expose insufficient coverage of creative workflows(Table \ref{app-category}). The model performs well on office applications and system-level tasks, suggesting macOS-focused training generalizes across typical desktop environments but requires targeted data collection for specialized professional domains. This can be mitigated by extending crawling to more creative apps and using accounts with pre-filled user-generated content in the future work, that allow for more content manipulation. 

\subsubsection{Agentic}\label{appendix:eval-agentic}
\begin{table}[!ht]
\centering
\renewcommand{\arraystretch}{1.3}
\begin{tabular}{lccc} 
\toprule
 & \multicolumn{3}{c}{\textbf{Success Rate (\%)}} \\ 
\textbf{Model} & Input & Click  & Overall \\ 
\midrule
OpenAI Computer Use    & 8.04 & \textbf{68.75}  & \textbf{64.41} \\ 
Claude Computer Use    & 8.93 & 65.59 & 61.53  \\ 
OS-Atlas-Pro-7B & 7.14 & 62.84   & 58.85   \\ 
UI TARS 1.5 7B & 1.79 & 54.65 & 50.86  \\ 
UI TARS 2B & 7.14 & 50.24  & 47.16 \\ 
Qwen 2.5 VL 3B & \textbf{12.5} & 42.95 & 40.77 \\ 
Qwen 2.5 VL 7B & 2.68 & 39.16 & 36.55  \\ 
CogAgent 9B & 3.57 & 15.83  & 14.95 \\ 
\bottomrule
\end{tabular}
\vspace{0.1in}
\caption{Success rate across models and interaction categories on \textsc{GUIrilla-Task} (agentic)}
\label{tab:success-rate-agentic}
\end{table}

\subsection{Impact of Backbone Model}
We further examined how the choice of backbone model influences performance. When trained on our dataset, a Qwen2-VL 7B backbone already surpasses OS-Atlas, despite both using the same underlying model. Notably, OS-Atlas was trained on nearly \textbf{300×} more data, yet our model achieves higher accuracy: +1.34\% in average and +2.02\% on the macOS subset of ScreenSpot-Pro. These results highlight the \textit{data efficiency} of our approach.

\newpage
\subsection{Ablation Details}\label{appendix:ablation}

\subsubsection{Influence of Handlers}\label{appendix:ablation-handlers}

\begin{table}[h!]
\centering
\small
\begin{tabular}{llccc}
\toprule
\textbf{Application} & \textbf{Metric} & \textbf{Handler-Supported Crawler} & \textbf{Random Crawler} \\
\midrule
\multirow{3}{*}{Stocks} 
& Tree Depth               & 14            & 16              \\
& Number of Tasks          & 162           & 32              \\
& Duplicate rate           & 0.08          & 0.14             \\
& Parse Time (hh:mm:ss)    & 00:20:51      & 00:29:35         \\
\midrule
\multirow{3}{*}{Maps} 
& Tree Depth               & 6            & 9              \\
& Number of Tasks          & 107           & 36              \\
& Duplicate rate           & 0.1          & 0.2             \\
& Parse Time (hh:mm:ss)    & 00:21:00      & 00:25:35         \\
\midrule
\multirow{3}{*}{Weather} 
& Tree Depth               & 6            & 7              \\
& Number of Tasks          & 73           & 73              \\
& Duplicate rate           & 0.0           & 0.01             \\
& Parse Time (hh:mm:ss)    & 00:10:56      & 01:05:48         \\
\bottomrule
\end{tabular}
\vspace{0.1in}
\caption{Comparison of Handler-Supported Crawler vs Random Crawler Across Applications}
\label{tab:ablation-handlers}
\end{table}

\subsubsection{Comparing Deterministic and GPT-Refined Task Descriptions}\label{appendix:gpt4-support}

\begin{table}[h]
\centering
\small
\caption{Examples of Task Agent and Deterministic Instructions by App}
\begin{tabular}{l p{5.2cm} p{6cm}}
\toprule
\textbf{App} & \textbf{Task Agent} & \textbf{Deterministic} \\
\midrule
Prayer Notes & Access Prayer Notes support page & button \\
\midrule
GoProPlayer & Open a media file & button Open\_Media… \\
\midrule
Fax & Add files or images to the fax & ADD FILES OR IMAGES button \\
\bottomrule
\end{tabular}
\label{tab:task_examples-ablation}
\end{table}

\subsection{\textsc{GUIrilla-Gold} Dataset Annotation Guidelines}\label{app:annotator-guidelines}

\subsubsection{Overview}
To assess the reliability of macOS accessibility (AX) metadata and the quality of GPT-generated task strings, we hired 5 annotators, who were given the test split data. Annotators with accessibility expertise reviewed each data entry along five dimensions: (1) task feasibility; (2) task instruction clarity and editing for ambiguity; (3) manual task execution; (4) accessibility tree quality rating (Good/Medium/Bad scale); and (5) element-level verification of semantic properties (role, description, value) and bounding-box accuracy.

\textbf{Task Quality}. From the 1319 original English language-based tasks, 84.3\% of tasks were marked as DOABLE after manual verification. Comparing GPT strings to human edits, 91\% required no change. The 109 edited cases showed 97\% similarity to originals (Ratcliff/Obershelp), confirming minor edits. We release manully edited dataset as \textsc{GUIrilla-Gold} \footnote{\url{https://huggingface.co/datasets/GUIrilla/GUIrilla-Gold/}}.

\textbf{AX Quality}. 
Accessibility metadata quality varies significantly: 64\% of screens received GOOD ratings, 24\% MEDIUM, and 12\% BAD. At the element level, only 40\% have correct role and description pairs, while 49\% contain role information only, and 11\% are mislabeled. Bounding boxes are accurate for 80\% of elements, though 10\% extend outside the visible window. This metadata sparsity and noise make accessibility-only task generation unreliable. We therefore recommend combining accessibility trees with screenshots and applying vision-based semantic adjustment to generate more precise, function-oriented, visually grounded tasks.

This document provides instructions for annotators to evaluate and improve UI task datasets, with focus on accessibility principles.

\subsubsection{Task String Feasibility Evaluation}
\textbf{Evaluation Steps:}
\begin{itemize}
    \item \textbf{Step 1:} Evaluate clarity and readability of the task string. Edit if ambiguous or poorly phrased.
    \item \textbf{Step 2:} Assess executability. Mark as DOABLE if the task is clear and the required element is visible. Mark as NOT DOABLE if the element is not visible or does not exist.
\end{itemize}

\textbf{DOABLE Examples:} ``Click the Submit button'', ``Type `hello world' in the search field''\\
\textbf{NOT DOABLE Examples:} ``Click the button'' (when multiple buttons present), ``Enter your password'' (if no password input visible)

\subsubsection{Task Execution Guidelines}
Attempt to execute the task exactly once to verify correctness:
\begin{itemize}
    \item \textbf{Click Actions:} Locate the correct element and click once within its bounding box
    \item \textbf{Type Actions:} Find the input field and type the exact text provided (case-sensitive)  
    \item \textbf{Multi-step Tasks:} Mark as NOT DOABLE if requiring multiple distinct actions
\end{itemize}

\textbf{Constraints:} Attempt only once, no retries. Do not fabricate actions not in the task string.

\subsubsection{Accessibility Quality Rating (1--3 Scale)}
\textbf{Score 1 -- BAD:} Critical issues severely impact assistive technology---missing labels, incorrect roles, invisible elements, broken grouping, no logical structure.

\textbf{Score 2 -- MEDIUM:} Moderate issues present---incomplete/generic labels, occasional role mismatches, partial grouping, minor positioning issues.

\textbf{Score 3 -- GOOD:} Accessibility tree accurately represents visual UI---descriptive labels, accurate roles, proper grouping, logical structure, complete state information.

\subsubsection{Label and Role Verification}
Review each element's semantic description and role. Uncheck ``Semantic'' if the meaning or AX role is incorrect. Uncheck ``BBox'' if the bounding box doesn't match the visible element.

\subsubsection{Quality Assurance Checklist}
\begin{itemize}
    \item[$\square$] Task string evaluated and edited for clarity
    \item[$\square$] NOT DOABLE marked for invisible elements  
    \item[$\square$] Task execution attempted once
    \item[$\square$] Accessibility score reflects usability
    \item[$\square$] Labels and roles verified
    \item[$\square$] Checkboxes unchecked for mismatches
\end{itemize}

\subsection{Confidence Intervals and Sensitivity to Decoding Strategies}\label{app:confidence-intervals}

To provide uncertainty estimates and strengthen the reliability of model comparisons, we conducted additional experiments examining performance variability across different decoding strategies.

\subsubsection{Experimental Setup}
Our primary results were obtained using greedy decoding with the following generation parameters: \texttt{num\_beams=3}, \texttt{do\_sample=False}, \texttt{temperature=None}, \texttt{top\_p=None}, \texttt{top\_k=None}.

To assess performance variability, we re-evaluated all models using stochastic decoding with parameters: \texttt{num\_beams=3}, \texttt{do\_sample=True}, \texttt{temperature=0.3}. For each model, we conducted three independent runs and computed mean ± standard deviation of success rates.

\subsubsection{Results}

Table~\ref{tab:confidence-intervals} presents results for both decoding strategies on the GUIrilla-Task test set.

The results reveal distinct patterns in decoding sensitivity. Fine-tuned GUIrilla-See models demonstrate remarkable consistency across decoding strategies, with standard deviations below 0.32\% in all cases. This stability suggests robust learning of UI interaction patterns.

In contrast, several pretrained models show significant performance degradation under stochastic decoding, most notably UI-TARS models which experience drops of 16-29 percentage points. This sensitivity highlights the importance of decoding strategy selection and suggests that some models may be overfitting to specific generation patterns during pretraining.

The minimal variance observed in our fine-tuned models provides confidence in the reported performance gains and demonstrates the robustness of the GUIrilla training approach across different inference conditions.

\begin{table}[t!]

\centering
\begin{tabular}{lcc}
\toprule
\textbf{Model} & \textbf{Greedy (\%)} & \textbf{Sampled (T=0.3) (\%)} \\
\midrule
Florence Base & 10.73 & 10.64 ± 0.10 \\
Florence Large & 8.31 & 8.08 ± 0.10 \\
Qwen 2.5 VL 3B & 17.96 & 18.06 ± 0.22 \\
Qwen 2.5 VL 7B & 35.78 & 35.40 ± 0.45 \\
UI-TARS 2B & 47.54 & 18.43 ± 0.22 \\
UI-TARS 1.5 7B & 69.07 & 52.17 ± 0.42 \\
UGround v1 2B & 64.03 & 64.41 ± 0.00 \\
UGround v1 7B & 69.46 & 69.84 ± 0.06 \\
OS-Atlas-Base-7B & 64.86 & 61.82 ± 0.10 \\
\midrule
GUIrilla-See-0.7B & 53.55 & 53.48 ± 0.32 \\
GUIrilla-See-3B & 73.48 & 73.90 ± 0.03 \\
GUIrilla-See-7B & 75.59 & 75.85 ± 0.06 \\
\bottomrule
\end{tabular}
\caption{Performance comparison between greedy and stochastic decoding strategies. Models fine-tuned on GUIrilla-Task (bottom section) show consistent performance with minimal variance, while some pretrained models exhibit notable degradation under stochastic decoding.}
\label{tab:confidence-intervals}
\end{table}
\end{document}